\documentclass[conference]{IEEEtran}
\IEEEoverridecommandlockouts
\usepackage{cite}
\usepackage{amsmath,amssymb,amsfonts,amsthm}
\usepackage{algorithmic}
\usepackage{graphicx}
\usepackage{textcomp}
\usepackage{xcolor}
\usepackage[ruled, linesnumbered]{algorithm2e}
\usepackage{hyperref}
\usepackage{booktabs, ragged2e}
\usepackage{multirow}
\usepackage[flushleft]{threeparttable}
\usepackage{makecell}
\usepackage{pifont}
\usepackage{array}
\usepackage{pifont}
\usepackage{enumitem}

\def\BibTeX{{\rm B\kern-.05em{\sc i\kern-.025em b}\kern-.08em
    T\kern-.1667em\lower.7ex\hbox{E}\kern-.125emX}}

\DeclareRobustCommand*{\IEEEauthorrefmark}[1]{\raisebox{0pt}[0pt][0pt]{\textsuperscript{\footnotesize #1}}}

\begin{document}

\title{FedCAP: Robust Federated Learning via Customized Aggregation and Personalization}

\author{
\IEEEauthorblockN{
Youpeng Li\IEEEauthorrefmark{1}, 
Xinda Wang\IEEEauthorrefmark{1}, 
Fuxun Yu\IEEEauthorrefmark{2}, 
Lichao Sun\IEEEauthorrefmark{3}, 
Wenbin Zhang\IEEEauthorrefmark{4},
Xuyu Wang\IEEEauthorrefmark{4}\IEEEauthorrefmark{*}
\thanks{\textsuperscript{*}Corresponding author}}
\IEEEauthorblockA{
\IEEEauthorrefmark{1}University of Texas at Dallas, Richardson, United States \\
\IEEEauthorrefmark{2}Microsoft, Redmond, United States \\ 
\IEEEauthorrefmark{3}Lehigh University, Bethlehem, United States \\
\IEEEauthorrefmark{4}Florida International University, Miami, United States \\
\href{mailto:youpeng.li@utdallas.edu,xinda.wang@utdallas.edu}{\texttt{\{youpeng.li,xinda.wang\}@utdallas.edu}}, \href{mailto:fuxunyu@microsoft.com}{\texttt{fuxunyu@microsoft.com}},\\ \href{mailto:lis221@lehigh.edu}{\texttt{lis221@lehigh.edu}}, \href{mailto:wenbin.zhang@fiu.edu,xuywang@fiu.edu}{\texttt{\{wenbin.zhang, xuywang\}@fiu.edu}}
}}

\maketitle

\begin{abstract}
Federated learning (FL), an emerging distributed machine learning paradigm, has been applied to various privacy-preserving scenarios. However, due to its distributed nature, FL faces two key issues: the non-independent and identical distribution (non-IID) of user data and vulnerability to Byzantine threats. To address these challenges, in this paper, we propose FedCAP, a robust FL framework against both data heterogeneity and Byzantine attacks. The core of FedCAP is a model update calibration mechanism to help a server capture the differences in the direction and magnitude of model updates among clients. Furthermore, we design a customized model aggregation rule that facilitates collaborative training among similar clients while accelerating the model deterioration of malicious clients. With a Euclidean norm-based anomaly detection mechanism, the server can quickly identify and permanently remove malicious clients. Moreover, the impact of data heterogeneity and Byzantine attacks can be further mitigated through personalization on the client side. We conduct extensive experiments, comparing multiple state-of-the-art baselines, to demonstrate that FedCAP performs well in several non-IID settings and shows strong robustness under a series of poisoning attacks.
\end{abstract}


\section{Introduction}
With the emergence of large foundation models~\cite{ChatGPT}, model performance increasingly relies on high-quality and high-volume data. In fields such as Internet of Things (IoT)~\cite{DBLP:conf/mobicom/LiZZC22, lim2020federated, wang2019adaptive} and healthcare~\cite{DBLP:conf/mobisys/OuyangXFCPLXZ023, wang2020csi}, user data often contains a large amount of sensitive information. Various privacy-preserving policies such as the General Data Protection Regulation (GDPR)~\cite{horvitz2015data, liu2022right} restrict the collection of user data by a central server. As an emerging distributed machine learning paradigm, federated learning (FL)~\cite{DBLP:conf/aistats/McMahanMRHA17} allows user data to remain local while coordinating clients to train a global model. Due to its distributed nature, FL faces two key issues. First, statistical heterogeneity exists in user data. In real-world FL applications, such as Google's next word prediction, training a single global model that caters to the individual needs of all users is challenging due to their diverse language habits and regional cultures~\cite{DBLP:journals/ftml/KairouzMABBBBCC21}. Second, FL systems are vulnerable to Byzantine threats~\cite{acsac1, acsac2}, with malicious clients uploading arbitrary model updates to the server, which can greatly degrade model performance on any test inputs (i.e., untargeted poisoning attack~\cite{DBLP:conf/uss/FangCJG20}).

To mitigate the impact of data heterogeneity, the concept of personalized FL is introduced~\cite{DBLP:conf/nips/SmithCST17}, where each client holds a personalized model to fit its own data distribution better. However, most personalized FL algorithms~\cite{DBLP:conf/icml/00050BS21, DBLP:conf/iclr/ChenC22, DBLP:conf/iclr/ZhangSFYA21} fail to adapt to non-independent and identically distributed (non-IID) settings. Therefore, we need to address \textbf{Challenge 1}:

\begin{itemize}[leftmargin=*] \item \textit{How to design a personalized FL framework that exhibits adaptiveness in various heterogeneous data settings?} \end{itemize}

To defend against poisoning attacks, existing robust FL methods adopt diverse strategies, with some focusing on the server side such as detection~\cite{DBLP:conf/ndss/CaoF0G21, DBLP:conf/kdd/ZhangCJG22} and robust aggregation~\cite{DBLP:conf/icml/YinCRB18, DBLP:conf/nips/BlanchardMGS17, DBLP:conf/icassp/SattlerMWS20, DBLP:signguard}, while others concentrate on the client side through personalization~\cite{DBLP:conf/icml/00050BS21}. However, the above robust FL methods are less effective against attacks in non-IID settings~\cite{bucket,GAS}, due to the difficulties in distinguishing malicious behavior from clients. This leads to varying degrees of aggregation knowledge loss while defending against attacks, which in turn results in model performance degradation. Moreover, in settings with strong attacks~\cite{LIE, Shejwalkar2021ManipulatingTB, DBLP:conf/uai/XieKG19}, malicious clients can camouflage themselves as benign, making it more difficult for robust FL methods to detect them %,
and causing further deterioration in the benign models. Therefore, it is imperative to tackle \textbf{Challenge 2}: 
%\vspace{-0.5mm}
\begin{itemize}[leftmargin=*] \item \textit{How can we design a Byzantine-robust FL framework that precisely distinguishes between benign and malicious clients in non-IID settings without causing a significant loss in model accuracy?} \end{itemize}

For real-world applications, a unified FL framework considering both challenges is needed, but few studies focus on this. Although Ditto~\cite{DBLP:conf/icml/00050BS21} mitigates the impact of data heterogeneity and attacks via personalization, one of its limitations is that it does not directly detect malicious clients and instead requires a trade-off between model utility and robustness. A naive strategy is to combine robust aggregation rules (AGRs) (e.g., Median, Trimmed Mean~\cite{DBLP:conf/icml/YinCRB18}, and ClusteredFL~\cite{DBLP:conf/icassp/SattlerMWS20}) with client personalization (e.g., Ditto), but their inherent limitations lead to combinations that fail to improve model performance (see empirical results in Fig.~\ref{fig:Ditto+}).

Motivated by the limitations of the above approaches, we propose FedCAP, a robust FL framework against both data heterogeneity and Byzantine attacks. FedCAP mainly includes four components. A model calibration mechanism helps distinguish malicious model updates from benign ones in non-IID settings. A customized aggregation rule can then facilitate collaboration among similar clients and accelerate malicious model deterioration. With the incorporation of an anomaly detection mechanism, the server is able to identify and permanently remove malicious clients. A personalized training module can further mitigate the impact of data heterogeneity and attacks, building on the relatively clean customized models. These components enable FedCAP to adapt to various non-IID settings and types of attacks.

FedCAP utilizes two key insights observed and analyzed through experiments in Section~\ref{sec:insights}. First, following the principle of "good becomes better and bad becomes worse", we can promote collaboration among similar clients in benign scenarios and inhibit cooperation between benign and malicious clients in attack scenarios. Second, we observe abnormal behavior from malicious clients when they intrude into FL training. Specifically, we observe that as the number of global rounds increases, the average Euclidean norm of the model updates from malicious clients gradually rises, thus degrading the model performance of benign clients.

Building upon the above insights, we design the customized aggregation rule to address \textbf{Challenge 1}. The server assigns each client a customized model that closely matches its data distribution, determined by the contributions from other clients. Specifically, we use the normalized cosine similarity of model updates among clients as aggregation weights to customize the models. By adjusting the scale factor of normalization, FedCAP achieves adaptation for various non-IID settings. To tackle \textbf{Challenge 2}, we propose the model update calibration mechanism and the anomaly detection mechanism based on the Euclidean norm of calibrated model updates. Specifically, by calibrating the uploaded model updates, FedCAP effectively captures differences in the magnitude and direction of model updates between benign and malicious clients. Combining this with the customized aggregation, FedCAP accelerates the model deterioration of malicious clients, leading to a significant increase in the Euclidean norm of their calibrated model updates. This triggers the anomaly detection mechanism, allowing the server to identify and remove malicious clients. Therefore, FedCAP excels in distinguishing between benign and malicious clients in non-IID settings.

Our main contributions are summarized as follows:
%%\vspace{-1mm}
\begin{itemize}[leftmargin=*]
	\item We propose FedCAP, a robust FL framework against both data heterogeneity and attacks, which adapts to various non-IID settings and different types of attacks. %We will publicly release our code after acceptance.%is publicly available at https:XXX. 
	\item We propose a model update calibration mechanism that excels in capturing the difference in the direction and magnitude of client model updates in non-IID settings.
	\item We design a customized model aggregation rule, which facilitates collaboration among similar clients while accelerating the model deterioration of malicious clients, helping the server identify and remove them permanently by triggering an anomaly detection mechanism based on the Euclidean norm of calibrated model updates.
	\item We perform extensive experiments indicating that the proposed FedCAP outperforms the state-of-the-art (SOTA) FL baselines in terms of model accuracy and robustness.
\end{itemize}

\noindent\textbf{Open science.} The source code and data artifacts of FedCAP have been open-sourced at \href{https://github.com/youpengl/FedCAP}{Github}. 

\section{Related Work}\label{sec:rw}
\noindent\textbf{Federated Learning with Non-IID Data}. FL can be broadly categorized into two types: single-model FL and multi-model FL. In single-model FL, clients collaboratively train a single global model~\cite{DBLP:conf/aistats/McMahanMRHA17}. Existing research primarily focuses on two key aspects: enhancing the generalizability of the global model~\cite{DBLP:conf/mlsys/LiSZSTS20, DBLP:conf/icml/KarimireddyKMRS20, DBLP:conf/iclr/ChenC22} and developing personalized FL algorithms to mitigate the impact of data heterogeneity~\cite{DBLP:conf/icml/00050BS21, DBLP:conf/iclr/ChenC22, DBLP:conf/iclr/ZhangSFYA21, sun1}. For example, FedRoD~\cite{DBLP:conf/iclr/ChenC22} improves the generalizability of the global model by using balanced softmax loss to mitigate the effect of label distribution skew. FedAvg-FT~\cite{DBLP:journals/corr/abs-2108-07313} treats clients' updated local models as personalized models and evaluates their performance. Ditto~\cite{DBLP:conf/icml/00050BS21} trains a personalized model for each client while locally updating the global model.

Given the limitation of generalizability of the single global model, the concept of multi-model FL, which refers to the server holding multiple models, is introduced. For example, in clustering-based FL algorithms~\cite{DBLP:conf/icassp/SattlerMWS20, DBLP:conf/ijcnn/BriggsFA20}, the server divides clients into multiple clusters, and within each cluster, the clients collaborate to train a group model. Unlike clustering, FedFomo~\cite{DBLP:conf/iclr/ZhangSFYA21} probabilistically selects batches of uploaded models and distributes them to clients, which each calculate model weights and perform weighted aggregation to obtain tailored models. However, sending multiple models to the clients in each communication round introduces high communication overhead. Compared to FedFomo, FedCAP performs customized aggregation on the server side without incurring any additional communication overhead.

%\vspace{1mm}
\noindent\textbf{Byzantine-robust Federated Learning}. Given the distributed nature of FL, it is vulnerable to Byzantine threats~\cite{DBLP:conf/uss/FangCJG20, sun2, sun3}. To defend against Byzantine attacks, a series of server-side AGRs built on top of averaged aggregation have been proposed (e.g., Krum~\cite{DBLP:conf/nips/BlanchardMGS17}, Multi-Krum~\cite{DBLP:conf/nips/BlanchardMGS17}, Median~\cite{DBLP:conf/icml/YinCRB18}, RFA~\cite{RFA}, Trimmed Mean~\cite{DBLP:conf/icml/YinCRB18}, etc.). Since these AGRs assume that all clients' data are IID, their robustness is less effective in non-IID settings. For non-IID defenses, %Cao et al.~\cite{DBLP:conf/ndss/CaoF0G21} proposed FLTrust, where 
in FLTrust~\cite{DBLP:conf/ndss/CaoF0G21}, the server filters or processes abnormal model updates by checking the magnitude and direction of the client-uploaded model updates. However, FLTrust assumes the server holds a clean dataset to boost trust, which violates FL's privacy principles. To mitigate gradient heterogeneity, Karimireddy et al.~\cite{bucket} suggested dividing the uploaded model updates into several buckets before aggregation, averaging $s$ model updates within each bucket, and then using AGRs to aggregate the updates across buckets. To address the issue of the curse of dimensionality~\cite{curse}, which enables malicious gradients to circumvent defenses that aggregate all honest gradients, %Liu et al.~\cite{GAS} proposed GAS. 
GAS~\cite{GAS} splits high-dimensional gradients into $p$ low-dimensional sub-vectors, scores them with a robust AGR, and aggregates the gradients identified as honest based on low gradient scores. 

For client-side defenses, %Li et al.~\cite{DBLP:conf/icml/00050BS21} proposed Ditto. B
by training a personalized model for each client with a regularizer controlling the distance between the personalized model and the global model, Ditto~\cite{DBLP:conf/icml/00050BS21} mitigates the impact of data heterogeneity and attacks to some extent. However, Ditto cannot generalize well across various non-IID settings and different types of attacks due to the difficulty in balancing client personalization and learning from global knowledge.

A naive combination of server-side AGRs and client personalization (e.g., Ditto) cannot fully mitigate the impact of data heterogeneity and attacks in non-IID settings. Compared to the aforementioned methods, FedCAP is proposed for ensuring unified robustness against both data heterogeneity and attacks in non-IID settings.

\section{Background and Motivation}\label{sec:bg_mtv}
\subsection{Federated Learning}
The original goal of FL is to maintain user data locally while coordinating clients to train a single global model. The vanilla FL algorithm (i.e., FedAvg~\cite{DBLP:conf/aistats/McMahanMRHA17}) consists of three steps: In each round $t$, the server distributes the global model $\boldsymbol{w}^t$ to participating clients, which then perform local training with their private data, uploading their updated models to the server. The server performs weighted averaging aggregation to get the updated global model $\boldsymbol{w}^{t+1}$. The optimization problem of FedAvg can be expressed as follows:

%\begin{setstrech}
%\setstretch{0.5}
% \vspace{-3mm}
\begin{equation}\label{eq:FedAvg}
\boldsymbol{w}_k^{t}=\arg \min _{\boldsymbol{w}} \mathcal{L}_k(\boldsymbol{w})\;\;(\text{initialized with }\boldsymbol{w}^{t}),
\end{equation}
% \vspace{-3mm}
%\end{setstrech}

\noindent where $\mathcal{L}_k=\frac{1}{|\mathcal{D}_k|}\sum_i\ell(\boldsymbol{x}_i,y_i;\boldsymbol{w}),\;\;\boldsymbol{w}^{t+1} \leftarrow \sum_{k=1}^N p_k\boldsymbol{w}_k^{t}.\nonumber$
\noindent Here, $\mathcal{S}_{(k\in S)}$ represents the set of clients, $N$ is the number of participants in each round, $\mathcal{D}_k$ denotes the training dataset of client $k$, $\ell$ is the loss function, $(\boldsymbol{x}_i,y_i)$ denotes a sample pair, and $p_k=\frac{|\mathcal{D}_k|}{|\mathcal{D}|}$ represents the aggregation weight assigned to client $k$. However, real-world user data often exhibits non-IID characteristics with various distribution skews~\cite{DBLP:journals/ftml/KairouzMABBBBCC21}, making it challenging for a single global model to effectively generalize across heterogeneous data.\\
\textit{ Personalized Federated Learning}. To tackle the challenge of data heterogeneity, several personalized FL algorithms have been proposed. The optimization objective of the personalized model can be formulated in a general form as:

%\begin{setstrech}
%\setstretch{0.5}
% \vspace{-2mm}
\begin{equation}\label{eq:PFL}
\boldsymbol{v}^{t+1}=\arg \min _{\boldsymbol{v}} \mathcal{L}(\boldsymbol{v})+\lambda \mathcal{R}(\boldsymbol{v} , \boldsymbol{w^*
}),
\end{equation}
% \vspace{-3mm}
%\end{setstrech}

\noindent where $\boldsymbol{v}$ is initialized with $\boldsymbol{v}^t$ and represents the personalized model, $\boldsymbol{w}^*$ denotes the global knowledge, such as the global model $\boldsymbol{w}^t$\cite{DBLP:conf/icml/00050BS21}, $\mathcal{R}$ denotes the regularizer, and $\lambda$ denotes the regularization factor controlling the extent to which the personalized model $\boldsymbol{v}$ references the global knowledge $\boldsymbol{w}^*$.

\subsection{Limitations of the SOTA}

\begin{figure}[h]\centering
%\vspace{-4mm}
\includegraphics[width=0.48\textwidth]{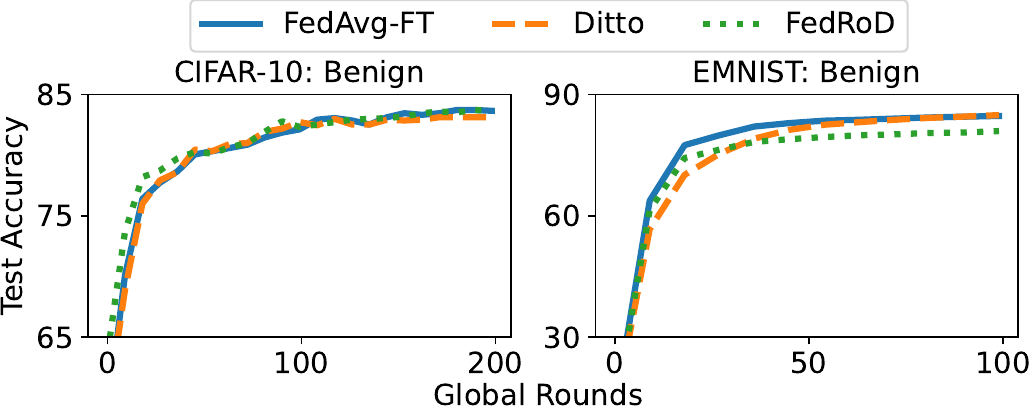}
% \vspace{-7mm}
\caption{Model performance comparison of SOTA FL methods
%, where PM and GM represent the personalized models and the global model, respectively.
} \label{fig:mtv-pfl}
\vspace{-4mm}
\end{figure}

\subsubsection{Limitations of Existing Personalized FL Methods}
To explore the limitations of representative FL approaches discussed in the related work, we evaluate the performance of FedAvg-FT~\cite{DBLP:journals/corr/abs-2108-07313}, Ditto~\cite{DBLP:conf/icml/00050BS21}, and FedRoD~\cite{DBLP:conf/iclr/ChenC22} in benign scenarios using both CIFAR-10~\cite{krizhevsky2009learning} and EMNIST~\cite{cohen2017emnist}. %, respectively.
For CIFAR-10, we adopt the pathological non-IID setting~\cite{DBLP:conf/aistats/McMahanMRHA17} to divide the data into 20 clients with a participating ratio of 1.0, where each client's data contains only two class types. For EMNIST, following a previous study~\cite{DBLP:conf/icml/KarimireddyKMRS20}, we distribute the data across 100 clients with a participation ratio of 0.2, allocating 20\% of the data as IID to each client and sorting the remaining 80\% based on labels.\footnote{If not specifically mentioned, these parameter settings are used by default for all experiments in Section~\ref{sec:bg_mtv}.} To simulate a realistic setting, the size of each client's sample is limited to a few hundred. We report the average test accuracy of personalized models.
From Fig.~\ref{fig:mtv-pfl}, we summarize as follows:
\begin{itemize}[leftmargin=*]
    \item In benign scenarios, the performance of the above personalized FL methods is comparable to or even worse than FedAvg-FT. This suggests that they are effective only in specific non-IID settings and struggle to adapt well to various non-IID settings with different distribution skew. Similar statements can be found in~\cite{DBLP:conf/iclr/ChenC22, DBLP:journals/corr/abs-2108-07313}.
\end{itemize}

\begin{figure}[h]\centering
  %\vspace{-3mm}
  \includegraphics[width=0.48\textwidth]{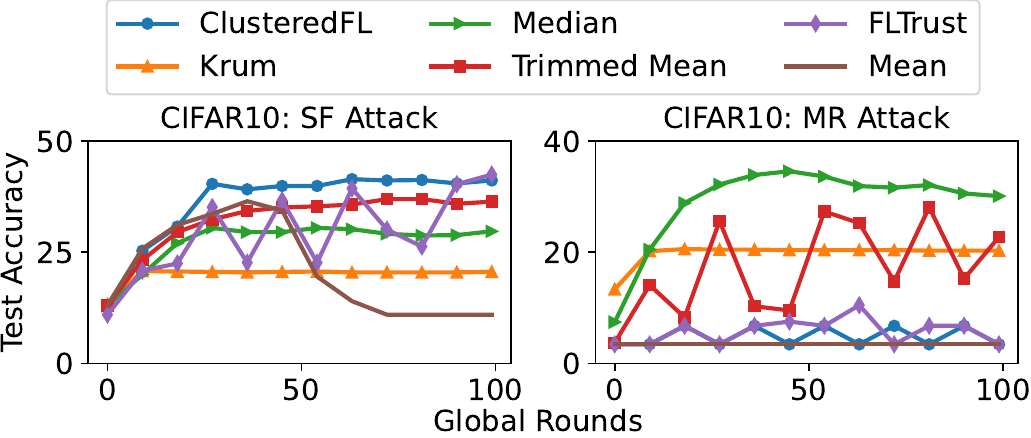}
% \vspace{-7mm}
\caption{Model performance comparison of robust FL methods} \label{fig:mtv-ram}
\vspace{-1mm}
\end{figure}

\subsubsection{Limitations of Existing Robust FL Methods}\label{sec:limitrobust}
To assess the robustness of existing robust FL methods, we evaluate the test accuracy of the global model on CIFAR-10~\cite{krizhevsky2009learning}. Specifically, we examine the performance of FedAvg~\cite{DBLP:conf/aistats/McMahanMRHA17}, FLTrust~\cite{DBLP:conf/ndss/CaoF0G21}, and AGRs including Krum~\cite{DBLP:conf/nips/BlanchardMGS17}, Median~\cite{DBLP:conf/icml/YinCRB18}, Trimmed Mean~\cite{DBLP:conf/icml/YinCRB18}, and ClusteredFL~\cite{DBLP:conf/icassp/SattlerMWS20} under Sign Flipping (SF) and Model Replacement (MR) attacks~\cite{Flip} detailed in Section~\ref{bases}. Fig.~\ref{fig:mtv-ram} shows that:
%\vspace{-0.5mm}
\begin{itemize}[leftmargin=*]
    \item Existing robust FL methods exhibit varying degrees of defense against attacks but often sacrifice valuable knowledge, as they struggle to distinguish malicious clients from benign ones in non-IID settings. Similar conclusions have been made in~\cite{GAS, bucket}.
\end{itemize}

\subsection{Insights and Motivations}\label{sec:insights}

\subsubsection{Good Becomes Better; Bad Becomes Worse}
Despite the statistical heterogeneity of data among clients, inherent similarities (e.g., common features) in data distributions still exist among some clients~\cite{DBLP:journals/sigkdd/KwapiszWM10}. Hence, promoting collaboration among similar clients can be advantageous. As the number of global rounds increases, models among similar clients become more similar and are less influenced by data heterogeneity. Moreover, by identifying anomalies in malicious behaviors, we can inhibit the cooperation between benign and malicious clients, safeguarding benign models from attacks. With more global rounds, malicious models are updated in a worse direction, accelerating their deterioration.
%\vspace{-1mm}

\subsubsection{Abnormal Euclidean Norm of Malicious Model Updates}

\begin{figure}[h]\centering
%\vspace{-4mm}  
\includegraphics[width=0.48\textwidth]{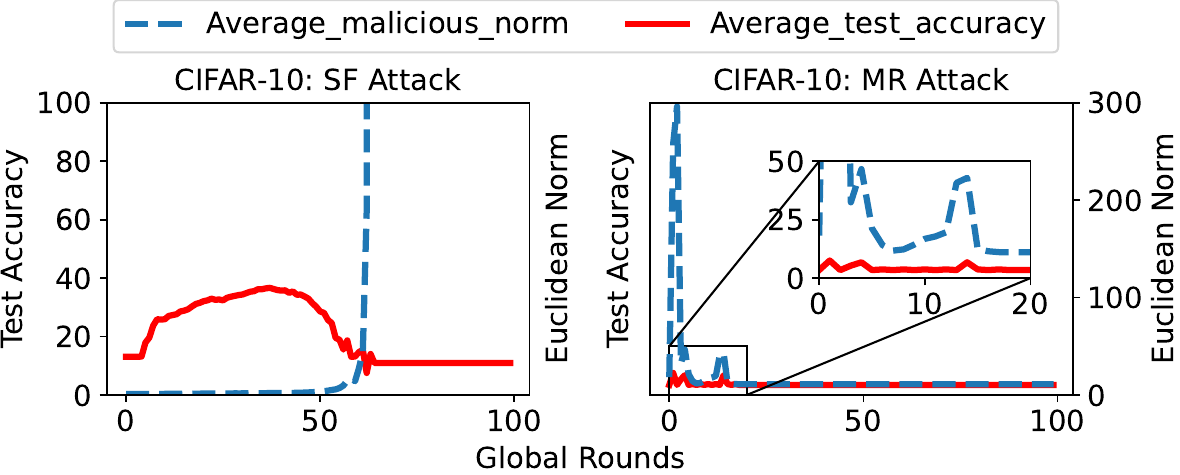}
% \vspace{-6mm}
\caption{Impact of model poisoning attacks on Euclidean norm of updates} \label{fig:mtv-norm}
\vspace{-4mm}
\end{figure}

To investigate the impact of Byzantine attacks on FL training, we evaluate FedAvg~\cite{DBLP:conf/aistats/McMahanMRHA17} under SF~\cite{Flip} and MR~\cite{DBLP:conf/icml/00050BS21} attacks using CIFAR-10~\cite{krizhevsky2009learning}. Fig.~\ref{fig:mtv-norm} illustrates that in both attack scenarios, the Euclidean norm of model updates uploaded by malicious clients increases dramatically as the number of global rounds increases, deepening the impact of attacks on the global model. Ultimately, FL training becomes dominated by attacks, resulting in a substantial decrease in the model performance of benign clients. 

These insights motivate us to propose the customized model aggregation rule, design the model update calibration mechanism, and develop the anomaly detection module. These innovations facilitate collaboration among similar clients, accurately capture differences in model updates between benign and malicious clients, and empower the server to identify and remove malicious clients.

\section{Problem Formulation}

\subsection{Aggregation Function}
In Eq.~\ref{eq:FedAvg}, the aggregation rule ignores the contributions of clients to each other, leading to the global model that unfairly favors clients with more samples. Our approach falls into the category of multi-model FL, where the server aggregates multiple models. Unlike Eq.~\ref{eq:FedAvg}, FedCAP customizes the aggregation weights for each client. The aggregation of client $k$'s customized model $\boldsymbol{\hat{w}}_k$ can be expressed as:

%\begin{setstrech}
%\setstretch{0.3}
\begin{equation}\label{eq:FedCAP}
    \boldsymbol{\hat{w}}_k \leftarrow \sum_{i=1}^N p_{ki}\boldsymbol{w}_i,
\end{equation}
%\end{setstrech}

\noindent where $p_{ki}$ denotes the contribution between client $k$ and $i$.
Eq.~\ref{eq:FedCAP} takes into account inter-client contributions, enabling the customized model of client $k$ to better match its data distribution. Further details are discussed in Section~\ref{sec:customization}.

% \vspace{-2mm}

\subsection{Threat Model}\label{threat}
% \vspace{-1mm}
In this work, our focus is on enhancing Byzantine-robustness in FL against poisoning attacks. 

\noindent\textbf{Adversary's Goal.} The objective of the attack is to disrupt the FL training process, resulting in a significant degradation in model performance on any test inputs (i.e., untargeted poisoning attack~\cite{DBLP:conf/uss/FangCJG20}). In real-world scenarios, such attacks could cause FL systems to crash, leading to inaccurate model inference in downstream tasks (e.g., disease diagnosis), which could result in immeasurable losses~\cite{kairouz2021advances}.

\noindent\textbf{Adversary's Capabilities.} Adversaries may intrude into FL systems by injecting fake clients or compromising benign ones. Considering the attack's cost and feasibility in real-world scenarios, the proportion of malicious clients typically does not exceed 50\%~\cite{DBLP:conf/uss/FangCJG20}. Given their known knowledge, adversaries can manipulate the FL training process by uploading arbitrary or finely crafted malicious model updates to the server, affecting the model aggregation.%, to the server, affecting the model aggregation.

\noindent\textbf{Adversary's Knowledge.} We consider attack scenarios where adversaries have partial knowledge, including model updates, local data, and local update rules from malicious clients. Despite its limited practical applicability, we further introduce a full-knowledge scenario to explore the upper bound of Byzantine robustness in our method. Under this assumption, adversaries have full knowledge of all clients, including benign ones, enabling them to design stronger and adaptive attacks.

\section{Design of FedCAP}\label{sec:FedCAP_Sys}

\begin{figure}[h]\centering
% \vspace{-4mm}
    \includegraphics[width=0.35\textwidth]{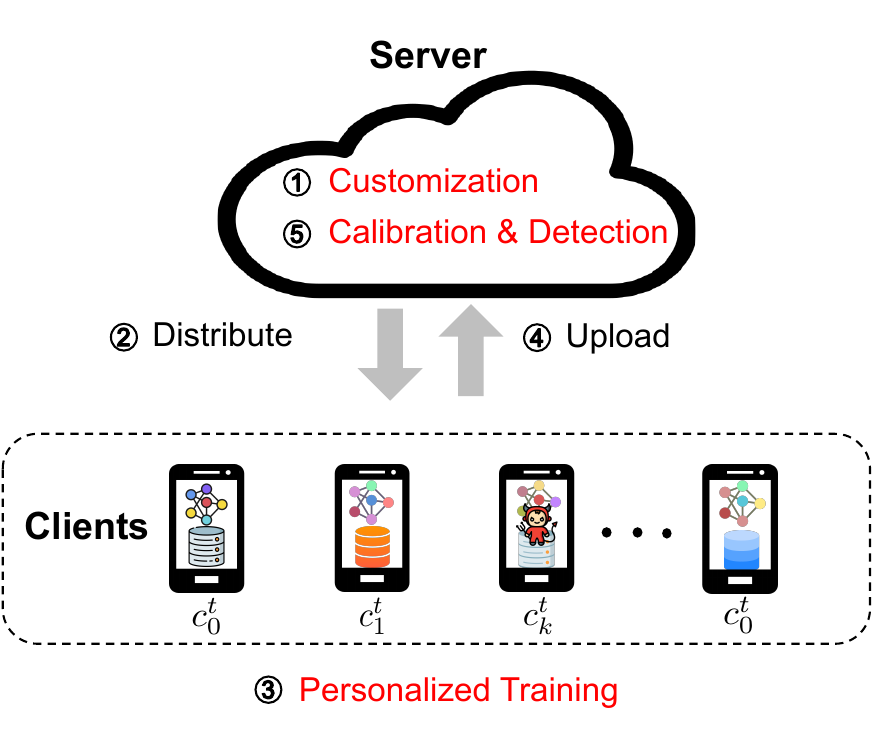}
\vspace{-2mm}
\caption{Workflow of FedCAP} \label{fig:Overview}
\vspace{-4mm}
\end{figure}

Fig.~\ref{fig:Overview} presents the workflow of FedCAP, comprising three main modules highlighted in red: model customization, personalized training, and model update calibration and detection. In each round, the server begins by customizing the models based on historical knowledge, which includes a recovered model pool and a calibrated update pool, along with the model updates uploaded by the clients (\ref{sec:customization}). Subsequently, these customized models are distributed to the participating clients. Upon receiving their respective customized models, the clients perform local updating and personalized training using their private data (\ref{sec:ptrain}). The server then calibrates the uploaded model updates to capture differences between clients, checks the calibrated model updates to detect malicious behavior, and updates historical knowledge (\ref{sec:HKU}). This iterative process continues for a certain number of rounds until the target accuracy is achieved or a pre-set number of rounds is reached.

\subsection{Model Customization}\label{sec:customization}

\begin{figure}[h]\centering
  \vspace{-2mm}
  \includegraphics[width=0.48\textwidth]{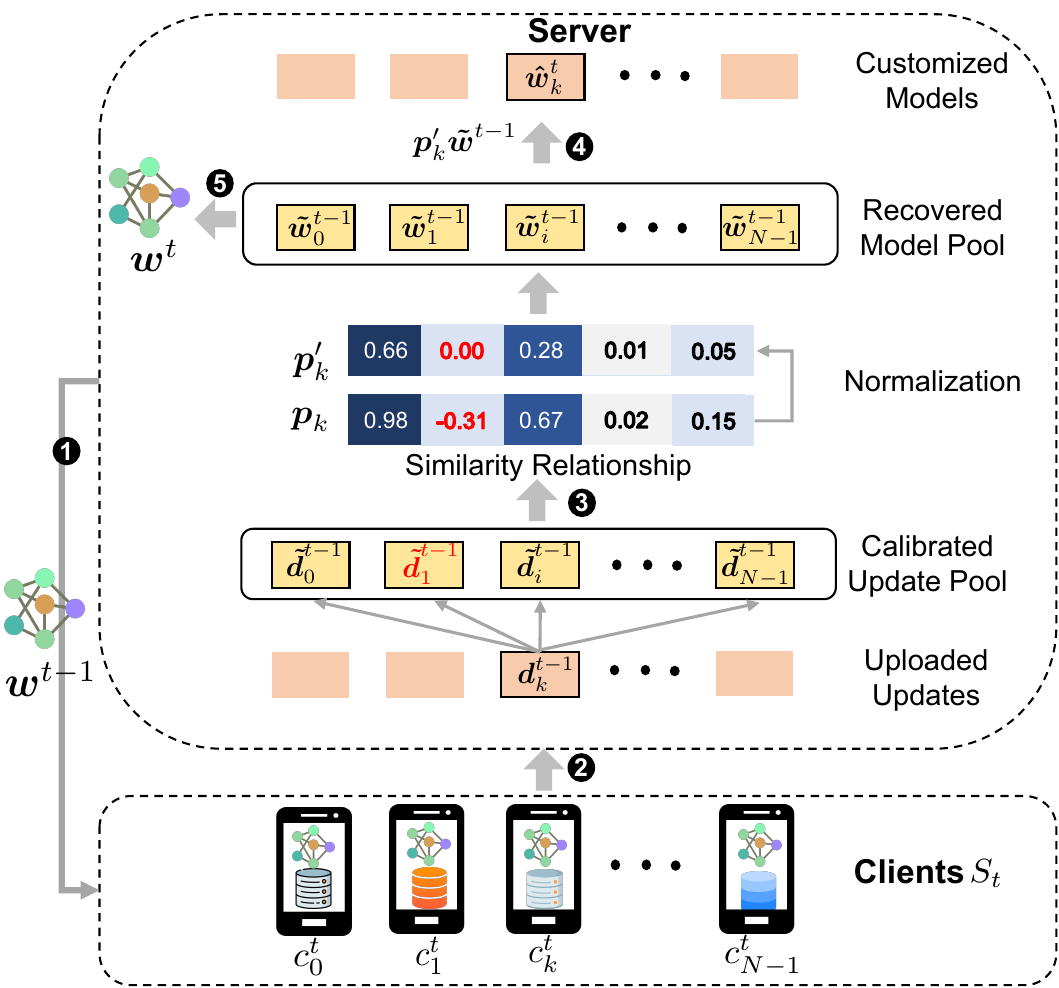}
% \vspace{-5mm}
\caption{Model customization} \label{fig:Customization}
\vspace{-1mm}
\end{figure}

We assume that FL training has progressed to round $t(t>0)$ and client $k$ is joining FL training, where $k\in S_t$ and $|S_t|=N$. The model customization for client $k$ includes the following steps:

%\vspace{1mm}
\noindent\textbf{Collection}. The server collects the model update $\boldsymbol{d}_k^{t-1}$ from client $k$ (as shown in \ding{182}-\ding{183} in Fig.~\ref{fig:Customization}). Depending on whether client $k$ participated in the previous round or is a new client, the server handles it differently:
\begin{itemize}[leftmargin=*]
    \item If client $k$ participated in round $t-1$ (i.e., $k\in S_{t-1}$), the server locally keeps the calibrated update pool $\{\boldsymbol{\tilde{d}}_i^{t-1}\}_{i\in S_{t-1}}$. In this case, there is no need to collect the model update $\boldsymbol{d}_k^{t-1}$ from client $k$ for round $t$ (i.e., $\boldsymbol{d}_k^{t-1}=\boldsymbol{\tilde{d}}_k^{t-1}$), which helps save communication overhead. Since we cannot directly calculate the contribution of client $k$ to itself, we define the contribution of client $k$ to itself (i.e., $p_{k,k}'$ in Eq.~\ref{eq:normalize}) as the weight factor $\phi$, and the remaining $1-\phi$ is assigned based on Eq.~\ref{eq:cosine} and Eq.~\ref{eq:normalize}. The impact of $\phi$'s value on model performance will be analyzed in Section~\ref{sec:phi}.
    \item If client $k$ did not join in round $t-1$ or is a new client, the server sends the global model $\boldsymbol{w}^{t-1}$ to client $k$, and the client returns the model update $\boldsymbol{d}_k^{t-1}$ to the server, where $\boldsymbol{d}^{t-1}_k=\boldsymbol{w}^{t-1}_k-\boldsymbol{w}^{t-1}$.
\end{itemize}

\noindent\textbf{Customized Aggregation}. Then, the server performs the customized model aggregation (as shown in \ding{184}-\ding{185} in Fig.~\ref{fig:Customization}). It first computes the cosine similarity (Eq.~\ref{eq:cosine}) between $\boldsymbol{d}_k^{t-1}$ and the calibrated update pool $\{\boldsymbol{\tilde{d}}_i^{t-1}\}_{i\in S_{t-1}}$:
%\vspace{-3mm}

%\begin{setstrech}
%\small
%\setstretch{0.5}
\begin{equation}\label{eq:cosine}
p_{k, i(k\neq i)}=\frac{<\boldsymbol{d}^{t-1}_k,\boldsymbol{\tilde{d}}^{t-1}_i>}{||\boldsymbol{d}^{t-1}_k||\cdot||\boldsymbol{\tilde{d}}^{t-1}_ i||},
\end{equation}
%\end{setstrech}

\noindent where the contribution $p_{k,i}$ between client $k$ and $i$ is determined by the similarity between calibrated model updates. The reason is that the similarity between model updates reflects the similarity of user data distribution~\cite{DBLP:conf/icassp/SattlerMWS20, DBLP:conf/ijcnn/BriggsFA20}. Clients with higher similarity contribute more valuable knowledge to each other. Therefore, when customizing aggregation weights, larger weights will be assigned to similar clients. In this way, the impact of data heterogeneity on customized models is mitigated by customized aggregation.

Since the cosine similarity takes values in the range of $[-1, 1]$, the softmax function is introduced for normalization (Eq.~\ref{eq:normalize}) to ensure that the sum of aggregation weights is $1$, and the weight of each client is non-negative.

%\begin{setstrech}
%\setstretch{0.5}
\begin{equation}\label{eq:normalize}
p_{k,i(k\neq i)}' = \frac{e^{\alpha p_{k, i}}}{\sum^{N}_{i}e^{\alpha p_{k,i}}}.
\end{equation}
%\end{setstrech}

\noindent The scale factor $\alpha$ in Eq.~\ref{eq:normalize} controls the sensitivity of the weight vector $\boldsymbol{p}'_k$ to the similarity vector $\boldsymbol{p}_k$. When $\alpha=0,\ p_{k,i}'=1/N$, and all participants have the same weight. Thus, by controlling the value of scaling factor $\alpha$, our customized aggregation can adapt to various degrees of data heterogeneity. Moreover, in attack scenarios, due to the low similarity between the model updates of malicious and benign clients, a large $\alpha$ is suggested to amplify the penalty for malicious clients, ensuring that the corresponding weights of the malicious clients after normalization converge to $0$. This prevents the aggregation process of the customized models of benign clients from being interfered with by malicious clients.

Once the aggregation weight vector $\boldsymbol{p}_k'$ is obtained, the server aggregates the recovered model pool $\{\boldsymbol{\tilde{w}}_i^{t-1}\}_{i\in S_{t-1}}$ based on $\boldsymbol{p}_k'$ to customize the model $\boldsymbol{\hat{w}}_k^t$ for client $k$ (Eq.~\ref{eq:FedCAP}).

\noindent\textbf{Global Model Updating}. At the same time, the server aggregates the recovered model pool $\{\boldsymbol{\tilde{w}}_i^{t-1}\}_{i\in S_{t-1}}$ to update the global model (as shown in \ding{186} in Fig.~\ref{fig:Customization}). Even though FedCAP provides the customized model for each client, it still aggregates the global model in each round for subsequent model update calibration (see Section~\ref{sec:HKU}). The aggregation of the global model is formulated as follows:

%\begin{setstrech}
%\small
%\setstretch{0.5}
\begin{equation}\label{eq:GMU}
\boldsymbol{w}^t\gets \sum_{i}^{S_{t-1}}\frac{\left|\mathcal{D}_i\right|}{|\mathcal{D}|}\boldsymbol{\tilde{w}}^{t-1}_i.
\end{equation}
%\end{setstrech}

\subsection{Personalized Training}\label{sec:ptrain}
%Compared
Unlike the global model aggregated in single-model FL, the customized model $\boldsymbol{\hat{w}}^t_k$ better matches the data distribution of client $k$. However, considering that the customized model may still be affected by slight data heterogeneity or attacks in situations where no explicit similarity relationship exists among clients, we additionally train a personalized model for each client to further mitigate the impact of heterogeneity or attacks. The optimization objective for personalized model of client $k$ is formulated as follows:

%\begin{setstrech}
% %\small
%\setstretch{0.5}
\begin{equation}\label{eq:FedCAP-PFL}
\boldsymbol{v}_k^{t+1}=\arg \min _{\boldsymbol{v}} \mathcal{L}_k(\boldsymbol{v})+\frac{\lambda}{2}\|\boldsymbol{v}-\boldsymbol{\hat{w}}^t_k\|_2^2,
\end{equation}
%\end{setstrech}

\noindent where $\boldsymbol{v}$ is initialized with $\boldsymbol{v}_k^{t}$. Eq.~\ref{eq:FedCAP-PFL} follows the general form of objective functions for personalized FL (Eq.~\ref{eq:PFL}). Distinguishing from the global model $\boldsymbol{w}^t$, here, $\boldsymbol{w^*}$ in Eq.~\ref{eq:PFL} denotes the customized model $\boldsymbol{\hat{w}}^t_k$, and the regularizer $\mathcal{R}$ uses the square of the L2-norm. The regularization factor $\lambda$ controls the extent to which client $k$'s personalized model $\boldsymbol{v}$ references the customized model $\boldsymbol{\hat{w}}^t_k$, further mitigating the impact of data heterogeneity or attacks.

In our implementation, client $k$ iteratively updates the personalized model $\boldsymbol{v}^t_k$ and the customized model $\boldsymbol{\hat{w}}^t_k$. When updating $\boldsymbol{v}^t_k$ with Eq.~\ref{eq:FedCAP-PFL}, the parameters of $\boldsymbol{\hat{w}}^t_k$ are frozen. Then, $\boldsymbol{\hat{w}}^t_k$ is updated in the same mini-batch SGD: $\boldsymbol{w}_k^{t}=\arg \min\limits_{\boldsymbol{w}} \mathcal{L}_k(\boldsymbol{w})$,
where $\boldsymbol{w}$ is initialized with $\boldsymbol{\hat{w}}^t_k$.

%\vspace{-2mm}
\subsection{Model Update Calibration and Detection}\label{sec:HKU}

\begin{figure}[h]\centering
  % \vspace{-4mm}
  \includegraphics[width=0.405\textwidth]{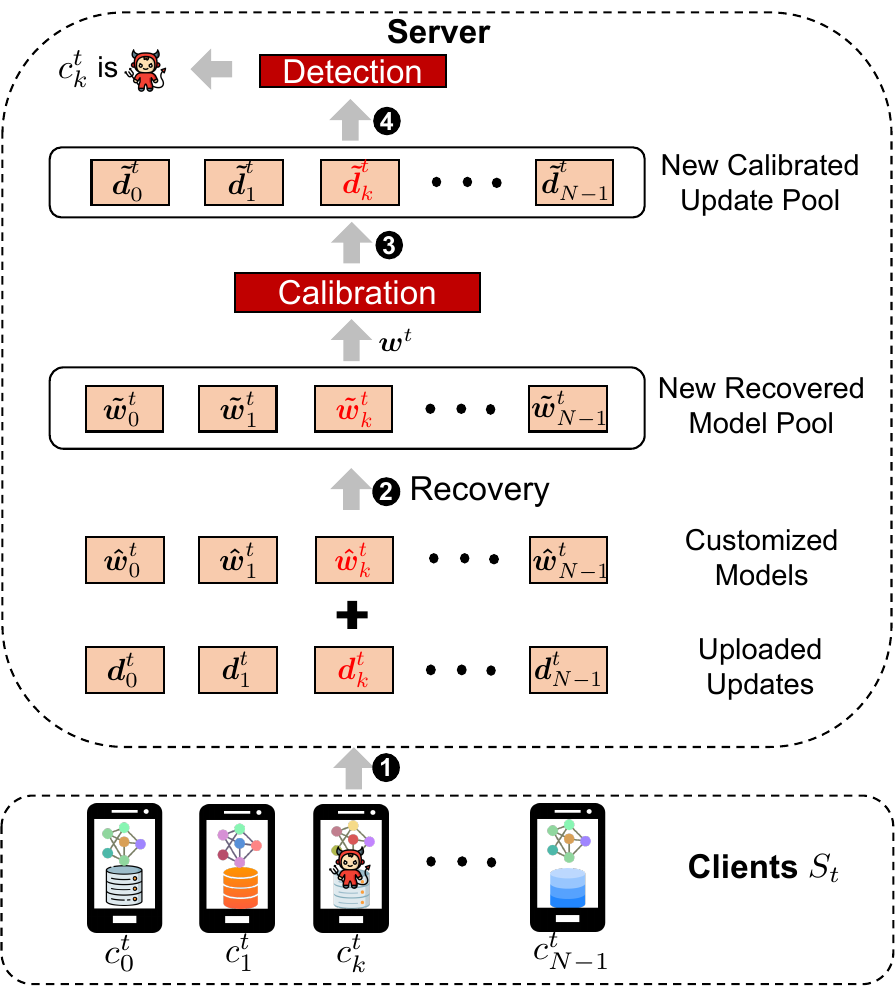}
% \vspace{1mm}
\caption{Model update calibration and detection} \label{fig:update}
\vspace{2mm}
\end{figure}

After client $k$ finishes updating the customized model, it returns the model update $\boldsymbol{d}^{t}_k$ to the server, where $\boldsymbol{d}^{t}_k=\boldsymbol{w}^{t}_k-\boldsymbol{\hat{w}}_k^t$ (as shown in \ding{182} in Fig.~\ref{fig:update}). In FedCAP, since the server customizes a unique model for each participating client, the starting points of local model updating in each round differ among clients\footnote{In FedAvg, the starting points of local model updating in each round are the same global model for all clients.}, making it challenging for the server to directly use the uploaded model updates $\{\boldsymbol{d}^t_k\}_{k\in S_t}$ to measure similarity relationships among clients during customized aggregation. To eliminate this inconsistency, we propose the model update calibration mechanism (as shown in \ding{184} in Fig.~\ref{fig:update}) that leverages the global model aggregated in each round (see Eq.~\ref{eq:GMU} in Section~\ref{sec:customization} for details) as a common reference point to calibrate the model updates uploaded by the clients.

\noindent\textbf{Recovery}. Before calibration, the server needs to recover the locally updated model $\boldsymbol{\tilde{w}}_k^t$ of client $k$ based on the customized model $\boldsymbol{\hat{w}}^t_k$ and the uploaded model update $\boldsymbol{d}^t_k$ (as shown in \ding{183} in Fig.~\ref{fig:update}), which can be formulated as $\boldsymbol{\tilde{w}}_k^{t}=\boldsymbol{\hat{w}}^t_k+\boldsymbol{d}^t_k$.

\noindent\textbf{Calibration}. The calibration process of the model update $\boldsymbol{d}^t_k$ uploaded by client $k$ can be represented as 
$\boldsymbol{\tilde{d}}^t_k=\boldsymbol{\tilde{w}}^t_k-\boldsymbol{w}^t$. After calibration, the server can not only accurately capture the similarity relationships among benign clients, but it can also differentiate malicious model updates from benign ones in non-IID settings. Here are some insights into the functionality of calibration in attack scenarios.

Specifically, we find that cosine similarity can only measure the directional differences of the uploaded model updates, which is why FLTrust~\cite{DBLP:conf/ndss/CaoF0G21} and ClusteredFL~\cite{DBLP:conf/icassp/SattlerMWS20} struggle to resist attacks targeting the magnitude of model updates (e.g., MR attack) in non-IID settings, as shown in Section~\ref{sec:limitrobust}. However, even though FedCAP also employs the same distance measurement criterion, it can effectively resist attacks targeting either the magnitude or direction of model updates. For example, when an adversary launches the model poisoning attack targeting the magnitude of the model update (e.g., MR attack), after calibration, the direction of the uploaded model update will be changed. This enables the distance measurement (i.e., cosine similarity) to capture the difference in calibrated model updates between benign and malicious clients.

\noindent\textbf{Detection}. For detection, the server calculates the Euclidean norm of the calibrated model update $\boldsymbol{\tilde{d}}^t_k$ for client $k$ (as shown in \ding{185} in Fig.~\ref{fig:update}). If the predetermined detection threshold $T_{norm}$ is exceeded, the server will recognize client $k$ as a malicious client and remove it permanently.

\setlength{\textfloatsep}{0pt}
\begin{algorithm}[htb]
%\setstretch{0.5}
\small
\caption{FedCAP}\label{alg:FedCAP}
\DontPrintSemicolon
\textbf{Input}: communication rounds $T$, client set $S$, initial global model $\boldsymbol{w}^0$ and personalized models $\{\boldsymbol{v}_k^0\}_{k\in S}$.\\
\textbf{Output}: model pool $\{\boldsymbol{\tilde{w}}_k^{T-1}\}_{k\in S_{T-1}}$, calibrated model update pool $\{\boldsymbol{\tilde{d}}_k^{T-1}\}_{k\in S_{T-1}}$, global model $\boldsymbol{w}^{T-1}$ and personalized models $\{\boldsymbol{v}_k^{T}\}_{k\in S}$.\\
\For{$t=0,\dots, T-1$}{
/* Server randomly selects a subset of clients $S_t$.*/\\
\uIf{$t\neq 0$}{
$\{\boldsymbol{\hat{w}}_k^t\}\gets$ Customize($\boldsymbol{w}^{t-1}, \{\boldsymbol{\tilde{d}}_i^{t-1}\}, \{\boldsymbol{\tilde{w}}_i^{t-1}\}$)\\
$\boldsymbol{w}^{t}\gets$ GlobalModelUpdating($\{\boldsymbol{\tilde{w}}_i^{t-1}\}$)
}
\Else{
/* Server initializes customized models*/\\
$\{\boldsymbol{\hat{w}}_k^0\}_{(k\in S_0)}$, where $\boldsymbol{\hat{w}}^0_k\gets \boldsymbol{w}^0$.
}
/* Server distributes $\{\boldsymbol{\hat{w}}_k^t\}_{(k\in S_t)}$ to clients $S_t$.*/\\
\For{each client $k \in S_t$}{
$\boldsymbol{w}_k^{t}, \boldsymbol{v}_k^{t+1}\gets$ ClientUpdate($\boldsymbol{\hat{w}}_k^t, \boldsymbol{v}_k^{t}$)\\
$\boldsymbol{d}_k^{t}\gets$Poison$(\boldsymbol{w}_k^{t}-\boldsymbol{\hat{w}}_k^t)$
}
/* Server calibrates uploaded model updates and detects whether client k is malicious.*/\\
$\{\boldsymbol{\tilde{w}}_k^{t}\}\gets$ Recover($\{\boldsymbol{\hat{w}}_k^t\},\{\boldsymbol{d}_k^{t}\}$)\\
$\{\boldsymbol{\tilde{d}}_k^{t}\}\gets$ Detect(Calibrate($\{\boldsymbol{\tilde{w}}_k^{t}\}, \boldsymbol{w}^t$))
}
% \vspace{-5mm}
\end{algorithm}
% \vspace{-2mm}
\subsection{Algorithm of FedCAP}
Alg.~\ref{alg:FedCAP} outlines the entire training process of FedCAP. When $t=0$, the server initializes the customized models $\{\boldsymbol{\hat{w}}^0_k\}_{k\in S_0}$ with $\boldsymbol{w}^0$ and distributes them to the corresponding participants $S_0$ (Line 9-12). Clients perform local updating to obtain updated local models $\{\boldsymbol{w}_k^{t}\}_{k\in S_t}$ and personalized models $\{\boldsymbol{v}_k^{t+1}\}_{k\in S_t}$ (Line 14). Subsequently, clients upload their model updates $\{\boldsymbol{d}_k^{t}\}_{k\in S_t}$ to the server (Line 15). The server recovers the updated models of clients, calibrates uploaded model updates, calculates the Euclidean norm of the calibrated model updates to detect whether the clients are malicious or benign, and updates the recovered model pool $\{\boldsymbol{\tilde{w}}^t_k\}_{k\in S_t}$ and the calibrated update pool $\{\boldsymbol{\tilde{d}}^t_k\}_{k\in S_t}$ (Line 18-19).

It's noteworthy that when $t=0$, all participants have identical parameters for the customized models $\{\boldsymbol{\hat{w}}^0_k\}_{k\in S_0}$. Therefore, the parameters of the calibrated model update of client $k$ are identical to those of its uploaded model update (i.e., $\boldsymbol{\tilde{d}}^t_k=\boldsymbol{d}^t_k$). When $t>0$, the server customizes models for clients through Section~\ref{sec:customization}. The subsequent steps remain consistent with those at $t=0$. Upon FL training, the server holds the recovered model pool $\{\boldsymbol{\tilde{w}}^{T-1}_k\}_{k\in S_{T-1}}$, the calibrated update pool $\{\boldsymbol{\tilde{d}}^{T-1}_k\}_{k\in S_{T-1}} $, and the global model $\boldsymbol{w}^{T-1}$.

\section{Experiment Evaluations}
% \vspace{-1mm}
\subsection{Experiment Setups}
% \vspace{-1mm}
\subsubsection{Datasets}
We use two image classification datasets, CIFAR-10~\cite{krizhevsky2009learning} and EMNIST~\cite{cohen2017emnist}, as well as a human activity recognition dataset, WISDM~\cite{DBLP:journals/sigkdd/KwapiszWM10}. 
The CIFAR-10 dataset contains images from 10 classes. We adopt a pathological non-IID setting~\cite{DBLP:conf/aistats/McMahanMRHA17} that introduces label distribution skew. Specifically, we define 20 clients, each with a balanced number of samples but imbalanced classes (2 classes per client).
%Regarding EMNIST, the original 
The EMNIST dataset comprises images of 62 different digits and letters. Following previous studies~\cite{DBLP:conf/icml/KarimireddyKMRS20, huang2021personalized}, we employ a non-IID setting to divide the data (commonly seen in cross-silo settings). Specifically, we define 100 clients and allocate data based on digits, lowercase letters, and uppercase letters, forming 3 distinct groups. Within each group, client data consists of 80\% samples from dominant classes and 20\% samples from all classes, leading to imbalanced sample distributions among the groups.
%Concerning WISDM, the original 
The WISDM data is collected from 36 user devices and 6 activity classes. We employ the default data distribution setting as the user-specific physiological and environmental variations introduce statistical heterogeneity (i.e., feature distribution skew) in the collected activity data.

For all datasets, we split the client data into training and test sets with a ratio of 0.75. To simulate scenarios where user sample sizes are limited, we restrict the number of samples for all clients to be on the order of hundreds.

\subsubsection{Models}
For all datasets, the models consist of two convolutional layers (with filter numbers ranging from 32 to 64 for CIFAR-10 and WISDM, and from 16 to 32 for EMNIST), followed by a fully connected layer (with 64 units for CIFAR-10 and WISDM, and 128 units for EMNIST), and an output layer. Note that considering the limited resources of user devices in real-world scenarios and the primary focus of this work on designing an FL algorithm, we do not explore other potentially better-performing models.

\subsubsection{Attack Methods}
Seven attacks are described below.

\noindent\textbf{Label Flipping (LF)}~\cite{Flip}: Malicious clients train models on manipulated data. The original data labels are flipped by $y_i'\gets(y_i+1)\%C$, where $C$ denotes the number of classes. 

\noindent\textbf{Sign Flipping (SF)}~\cite{Flip}: Malicious clients flip the signs of their model updates before uploading.

\noindent\textbf{Model Replacement (MR)}~\cite{DBLP:conf/icml/00050BS21}: Malicious clients scale up model updates by $N$ times before uploading. By default, $N$ is the number of participants per round. 

\noindent\textbf{A Little is Enough (LIE)}~\cite{LIE}: Malicious clients calculate mean and standard deviation for each coordinate over participant updates and set fake updates within the range of $(\mu_i-z^{max}\delta_i, \mu_i+z^{max}\delta_i)$, where $z^{max}$ is obtained from the Cumulative Standard Normal Function. 

\noindent\textbf{Min-Max}~\cite{Shejwalkar2021ManipulatingTB}: Min-Max attack optimizes a malicious gradient, ensuring the maximum distance between it and any benign gradient remains within the upper bound set by the largest distance between any two benign gradients. 

\noindent\textbf{Min-Sum}~\cite{Shejwalkar2021ManipulatingTB}: Min-Sum attack optimizes a malicious gradient, ensuring the sum of squared distances of the malicious gradient from all the benign gradients remains within the upper bound set by the sum of squared distances of any benign gradient from the other benign gradients. 
    The optimization problems for the Min-Max (Eq.~\ref{eq:min-max}) and Min-Sum (Eq.~\ref{eq:min-sum}) attacks can be formulated as follows:
    \vspace{-3mm}
    
    %\begin{setstrech}
    %\small
    %\setstretch{0.7}
    \begin{equation}
    \boldsymbol{d}_m=f_{\operatorname{avg}}\left(\boldsymbol{d}_{\{i \in[N]\}}\right)+\gamma \boldsymbol{d}_p,
    \end{equation}
    \vspace{-4mm}
    \begin{equation}\label{eq:min-max}
    \underset{\gamma}{\operatorname{argmax}} \max _{i \in[N]}\left\|\boldsymbol{d}_m-\boldsymbol{d}_i\right\|_2 \leq \max _{i, j \in[N]}\left\|\boldsymbol{d}_i-\boldsymbol{d}_j\right\|_2,
    \end{equation}
    \vspace{-2mm}
    \begin{equation}\label{eq:min-sum}
    \underset{\gamma}{\operatorname{argmax}} \sum_{i \in[N]}\left\|\boldsymbol{d}_m-\boldsymbol{d}_i\right\|_2^2 \leq \max _{i \in[N]} \sum_{j \in[N]}\left\|\boldsymbol{d}_i-\boldsymbol{d}_j\right\|_2^2, 
    \end{equation}
    %\end{setstrech}
    
    \noindent where $\boldsymbol{d}_m$ represents the malicious update, and $\boldsymbol{d}_p$ represents the perturbation vector. Following a previous study~\cite{Shejwalkar2021ManipulatingTB}, we choose $\boldsymbol{d}_p$ as -$std(\boldsymbol{d}_{\{i\in[N]\}})$.
    
    \noindent\textbf{Inner Product Manipulation (IPM)}~\cite{DBLP:conf/uai/XieKG19}: IPM attack aims to achieve a negative inner product between the true mean of the updates and the aggregation result, ensuring that the model update in the direction of gradient ascent.
    %\begin{setstrech}
    %\small
    %\setstretch{0.1}
    \begin{equation}
    \frac{1}{N} \sum_{i \in[N]} \Delta_i=\frac{N-M(1+\epsilon)}{N(N-M)} \sum_{m \in[M]} \Delta_m,
    \end{equation}
    %\end{setstrech}
    
    \noindent where $N$ denotes the number of participants, and $M$ denotes the number of malicious clients. To ensure that $\frac{N-M(1+\epsilon)}{N(N-M)}<0$, by default, we set $\epsilon$ to $N$~\cite{Blade}.

%(i.e., \textbf{Label Flipping (LF)}~\cite{Flip}, \textbf{Sign Flipping (SF)}~\cite{Flip}, \textbf{Model Replacement (MR)}~\cite{DBLP:conf/icml/00050BS21}, \textbf{A Little is Enough (LIE)}~\cite{LIE}, \textbf{Min-Max}~\cite{Shejwalkar2021ManipulatingTB}, \textbf{Min-Sum}~\cite{Shejwalkar2021ManipulatingTB}, and \textbf{Inner Product Manipulation (IPM)}~\cite{DBLP:conf/uai/XieKG19}), covering both data poisoning attacks and model poisoning attacks. (see Appendix~\ref{sec:atkm} for more details).
% \vspace{-1mm}
\subsubsection{Baselines}\label{bases}
To evaluate the effectiveness of personalized FL methods in various non-IID settings, we compare the performance of FedCAP with the six baselines: \textbf{Local Training}, \textbf{FedAvg}~\cite{DBLP:conf/aistats/McMahanMRHA17}, \textbf{FedAvg with Fine-tuning} (FedAvg-FT~\cite{DBLP:journals/corr/abs-2108-07313}),
\textbf{Ditto}~\cite{DBLP:conf/icml/00050BS21}, \textbf{FedRoD}~\cite{DBLP:conf/iclr/ChenC22}, and \textbf{FedFomo}~\cite{DBLP:conf/iclr/ZhangSFYA21} discussed in the related work.

To further validate the robustness of FedCAP, we compare it with \textbf{FLTrust}~\cite{DBLP:conf/ndss/CaoF0G21} and the following five AGRs. In \textbf{Multi-Krum (M-Krum)}~\cite{DBLP:conf/nips/BlanchardMGS17}, the server averages the top $Q$ models with the smallest scores to obtain the updated global model, with this work using $Q=N-M$. \textbf{Median}~\cite{DBLP:conf/icml/YinCRB18} takes the coordinate-wise median of the participants' client model parameters. \textbf{RFA}~\cite{RFA} takes the geometric-wise median of the participants' client model parameters. In \textbf{Trimmed Mean (Trim.)}~\cite{DBLP:conf/icml/YinCRB18}, the server sorts the model parameters of the participating clients by coordinates, then averages the remaining parameters after removing the Q largest and Q smallest values, where $Q=\lfloor\frac{M}{2}\rfloor$. Using \textbf{ClusteredFL}~\cite{DBLP:conf/icassp/SattlerMWS20}, the server identifies the optimal partitioning of clients into two clusters, and the cluster with the fewest clients is considered malicious. The server then aggregates the models within the remaining cluster. Considering some AGRs are IID defenses, we further combine them with two SOTA non-IID defenses (i.e., \textbf{GAS}~\cite{GAS} and \textbf{Bucketing}~\cite{bucket}) to explore their robustness in non-IID settings. 

\subsubsection{Parameter Settings}
Unless specified otherwise, in all experiments, we fix the batch size to 10, the learning rate to 0.01, global rounds to 100, and the epoch to 5. For CIFAR-10 and WISDM, the proportion of participating clients per round is set to 1.0, while for EMNIST, it is set to 0.2. In the attack scenarios, we assume a default proportion of malicious clients $p_{atk}$ of 0.3. In addition, we conduct a grid search for hyperparameters for all baselines. More specifically, for Ditto, the search range for $\lambda$ is $\{0.01, 0.1, 1, 2\}$. For FedFomo, we set the number of models sent to each client to half the proportion of participating clients~\cite{DBLP:conf/iclr/ZhangSFYA21}. For Bucketing, we set $s$ to 2 as per the original paper. For GAS, we search for the optimal number of sub-vectors $p$ in $\{1000, 10000, 100000\}$. As for FedCAP, the search range for $\lambda$ is $\{0.1, 0.5, 1\}$, for $\phi$ it is $\{0.1, 0.2, 0.3\}$, and for $\alpha$, it is $\{2, 5, 10\}$. In all attack scenarios, we fix the value of $T_{norm}$ to 10. This is because as the global rounds increase, the Euclidean norm of the calibrated model updates from malicious clients gradually approaches infinity, so the anomaly detection mechanism is not sensitive to $T_{norm}$.
% \vspace{-1mm}

\begin{figure*}[h]
\centering
  %\vspace{-4mm}
  \includegraphics[width=0.98\textwidth]{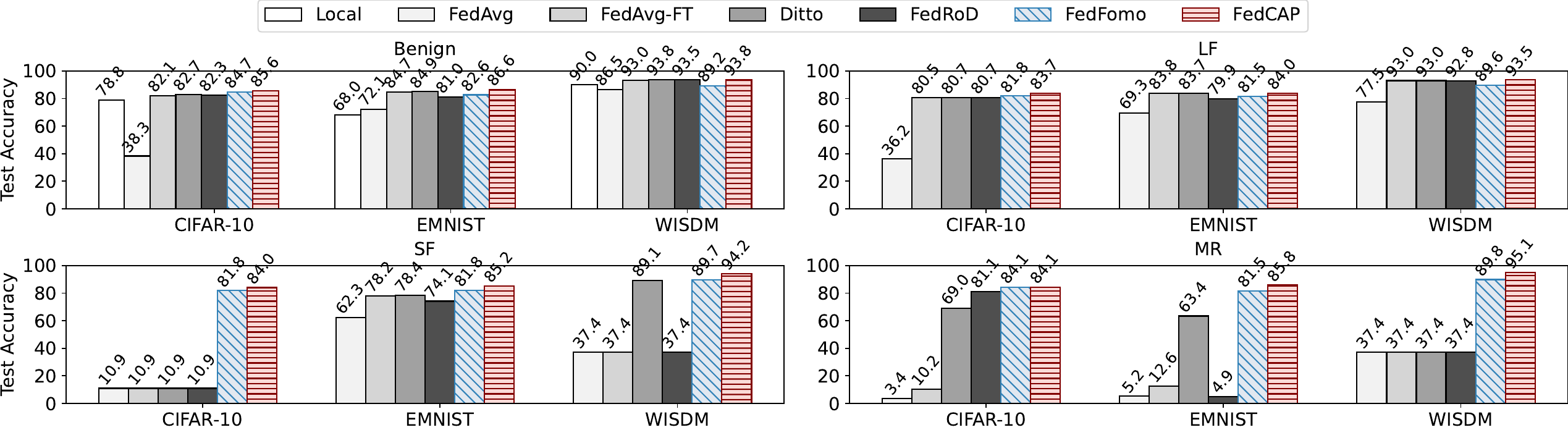}
  % \vspace{-2mm}
  \caption{Model performance comparison of FedCAP with SOTA FL baselines
}\label{fig:cpr-sota}
\end{figure*}

\subsubsection{Evaluation Metric}
In all experiments, we report the average \textbf{Test Accuracy (TAcc)} of all benign client models in the final global round. If not specifically mentioned, we choose the model with higher average test accuracy between customized models and personalized models.

\subsection{Performance Comparisons}
\subsubsection{Comparing with SOTA FL Baselines}\label{sec:cpr-sota}
We use the CIFAR-10, EMNIST, and WISDM datasets to compare the performance of FedCAP with other SOTA baselines in the benign scenario, and we also explore their potential defense ability in attack scenarios (i.e., LF, SF, and MR). In Fig.~\ref{fig:cpr-sota}, FedCAP outperforms all other baselines in all cases in terms of model accuracy. It exhibits an average accuracy gain of 2\% to 23\% over other baselines in the benign scenario and an average accuracy gain of 3\% to 50\% in attack scenarios. 
%Specifically, in the benign scenario with CIFAR-10 and EMNIST, FedCAP shows a significant performance gain compared to other baselines, which reveals clear similarity relationships among clients. Even on the relatively weakly heterogeneous WISDM dataset, FedCAP demonstrates comparable performance, indicating its adaptability to various degrees of data heterogeneity with the scale factor $\alpha$—10 for CIFAR-10 and EMNIST, and 2 for WISDM.

Specifically, in benign scenarios, although Ditto and FedRoD achieve comparable model performance on WISDM to FedCAP, their performance on CIFAR-10 and EMNIST shows a relatively marginal gap compared to FedCAP, averaging about 3\%. These results indicate that existing personalized FL algorithms, which rely solely on personalized training, cannot consistently achieve satisfactory performance under varying non-IID settings. In contrast, FedCAP adapts to different levels of data heterogeneity through customized aggregation (with the scale factor $\alpha$—10 for CIFAR-10 and EMNIST, and 2 for WISDM) and further mitigates the impact of data heterogeneity by incorporating the personalized training mechanism.

In attack scenarios, most methods fail to defend against attacks across heterogeneous datasets. Although FedFomo shows some robustness against attacks, balancing model utility and robustness through client-side personalized aggregation is challenging given the unknown number of malicious clients in real-world scenarios. Compared to others, FedCAP demonstrates strong robustness against all attacks. Especially under the MR and SF attacks, its model performance is on average about 3\% to 62\% higher than others. 

In summary, through the combined effects of customized aggregation and personalized training, FedCAP not only mitigates the impact of data heterogeneity, but it also effectively defends against malicious attacks by incorporating model update calibration and anomaly detection mechanisms. The above results show that FedCAP generalizes well to various data heterogeneity settings and attack scenarios.

\begin{figure*}[h]
\centering
  %\vspace{-4mm}
  \includegraphics[width=0.98\textwidth]{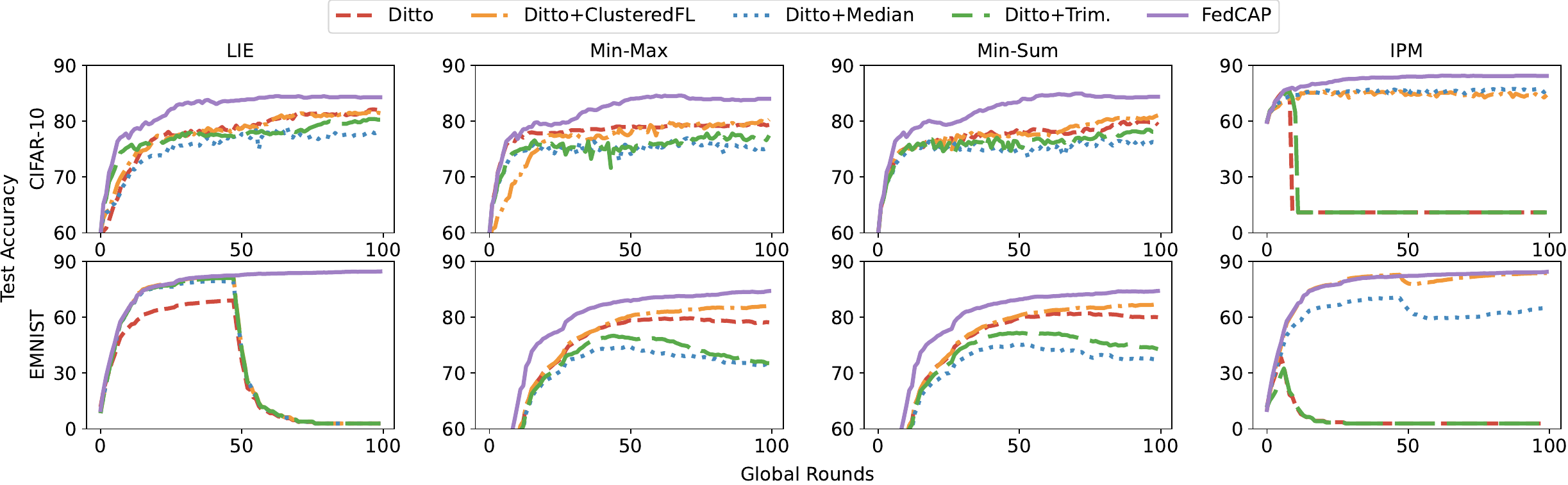}
  % \vspace{-3mm}
  \caption{AGRs augmented with Ditto
}\label{fig:Ditto+}
  \vspace{-3mm}
\end{figure*}

\subsubsection{Comparing with Robust FL Methods}\label{sec:cprram}To evaluate the robustness of FedCAP and existing robust FL methods, we use the CIFAR-10, EMNIST, and WISDM datasets to compare their model performances under six attack scenarios. For a fair comparison, we report the average test accuracy of their locally updated models. We do not consider the LF attack as it does not significantly affect most FL baselines in Section~\ref{sec:cpr-sota}. As can be seen from Table~\ref{tab:cpr-robust}, Median resists the attacks to some extent in most cases. ClusteredFL and FLTrust cannot counter the MR attack in non-IID settings because they use cosine similarity as the distance measurement, which cannot distinguish differences in the magnitude of model updates. Trimmed Mean fails to defend against the IPM attack. In contrast, FedCAP outperforms other robust FL methods in all cases with an average of 12\% to 27\% without a marginal accuracy loss compared to benign scenarios. Especially under the LIE attack on EMNIST, all methods except for FedCAP fail to resist the attack. This is attributed to the model update calibration in FedCAP, which helps capture the differences between malicious and benign model updates in non-IID settings, thereby aiding the server in reducing the impact of malicious model updates during customized aggregation.

In sum, the above results demonstrate that FedCAP can achieve superior robustness under non-IID settings and generalize to various attack scenarios.

\begin{table}[h]
  \caption{Model performance comparison of FedCAP with robust FL methods}
  \label{tab:cpr-robust}
  \centering
  \scriptsize
  \setlength{\tabcolsep}{0.5pt}
\begin{tabular}{c|c|cccccccc}
\hline
Dataset                   & Type    & Mean  & M-Krum & Median & RFA   & Trim. & ClusteredFL & FLTrust & FedCAP \\ \hline
\multirow{6}{*}{CIFAR-10} & SF      & 10.95 & 80.57  & 78.86  & 78.57 & 74.38 & 79.71       & 82.19   & \textbf{84.00} \\
                          & MR      & 10.19 & 82.38  & 82.38  & 82.38 & 81.24 & 17.24       & 20.67   & \textbf{84.10} \\
                          & LIE     & 80.86 & 79.90  & 77.24  & 78.00 & 80.19 & 82.57       & 72.10   & \textbf{84.95} \\
                          & Min-Max & 77.24 & 52.00  & 69.81  & 74.76 & 72.61 & 78.95       & 81.05   & \textbf{84.00} \\
                          & Min-Sum & 78.48 & 67.43  & 66.86  & 73.71 & 72.29 & 78.86       & 80.19   & \textbf{84.38} \\
                          & IPM     & 10.95 & 82.38  & 75.05  & 36.48 & 10.95 & 42.00       & 81.14   & \textbf{84.29} \\ \hline
\multirow{6}{*}{EMNIST}   & SF      & 78.15 & 81.80  & 79.55  & 80.27 & 79.66 & 80.04       & 84.05   & \textbf{85.24} \\
                          & MR      & 12.56 & 83.64  & 84.30  & 84.97 & 4.47  & 2.96        & 3.01    & \textbf{85.82} \\
                          & LIE     & 2.87  & 2.87   & 2.87   & 2.87  & 2.87  & 2.87        & 2.87    & \textbf{85.16} \\
                          & Min-Max & 76.86 & 71.08  & 70.31  & 71.55 & 69.59 & 81.28       & 83.01   & \textbf{85.39} \\
                          & Min-Sum & 78.76 & 71.68  & 71.06  & 71.80 & 72.02 & 81.61       & 83.37   & \textbf{85.58} \\
                          & IPM     & 2.87  & 2.06   & 65.88  & 19.06 & 2.87  & 83.72       & 84.13   & \textbf{85.43} \\ \hline
\multirow{6}{*}{WISDM}    & SF      & 37.36 & 90.23  & 92.25  & 91.04 & 89.53 & 89.02       & 93.35   & \textbf{94.16} \\
                          & MR      & 37.36 & 93.45  & 93.66  & 93.15 & 91.74 & 37.36       & 37.36   & \textbf{95.07} \\
                          & LIE     & 94.36 & 94.56  & 92.75  & 92.55 & 93.76 & 94.36       & 37.36   & \textbf{94.56} \\
                          & Min-Max & 90.13 & 92.35  & 91.44  & 92.65 & 82.78 & 91.84       & 91.54   & \textbf{94.06} \\
                          & Min-Sum & 90.23 & 93.45  & 91.34  & 92.75 & 83.69 & 93.35       & 92.15   & \textbf{94.26} \\
                          & IPM     & 37.36 & 93.35  & 81.67  & 82.38 & 37.36 & 37.36       & 93.25   & \textbf{94.06} \\ \hline
\end{tabular}
  \vspace{2mm}
\end{table}

\subsubsection{AGRs Augmented with Ditto}\label{sec:Ditto+}

% To further explore the effects of personalized FL augmented with AGRs, we combine Ditto, which is based on client-side personalization and defense, with three representative AGRs based on server-side defense. In Fig.~\ref{fig:Ditto+}, we show the convergence curves of personalized models, it can be seen that the incremental algorithms still cannot demonstrate reliable robustness against various attacks in the non-IID settings. In contrast, FedCAP not only demonstrates strong robustness in all cases but also converges faster than other methods. This is due to its approach of aggregating customized models based on the contributions (i.e., similarities) among clients, allowing client personalized models to reference more valuable information from the customized models.

Since FedCAP includes the server-side customized aggregation and the client-side personalization, we use the CIFAR-10 and EMNIST datasets to explore whether the combination of AGRs and SOTA personalized FL method Ditto is enough to excel in defending against attacks in non-IID settings. Fig.~\ref{fig:Ditto+} shows that although the combination of Ditto and ClusteredFL mitigates the impact of attacks such as IPM to some extent, its model performance is still on average about 13\% lower than that of FedCAP. Furthermore, in most cases, the model performance of Ditto combined with Median or Trimmed Mean even declines compared to Ditto. The reason is that these AGRs fail to detect malicious clients in non-IID settings, so the global model is still influenced by attacks, resulting in less valuable global knowledge for the personalized model to reference. 

In contrast, FedCAP not only demonstrates strong robustness in all cases but also converges faster than other methods. This is due to its model update calibration and anomaly detection mechanisms, which enable the server to detect malicious clients and aggregate customized models based on the contributions (i.e., similarities) among benign clients. This allows client personalized models to reference more valuable information from the customized models.
\vspace{-2mm}

\begin{table}[h]
%\vspace{-2mm}
  \caption{AGRs Augmented with Non-IID Denfenses}
  \label{tab:agraug}
  %\vspace{-3mm}
  \centering
  \scriptsize
  \setlength{\tabcolsep}{3pt}
\begin{tabular}{c|c|cc|cc|cc|c}
\hline
                         & Method & \multicolumn{2}{c|}{MKrum}                                                                                     & \multicolumn{2}{c|}{Median}                                                                                     & \multicolumn{2}{c|}{RFA}                                                                                        & \multirow{2}{*}{FedCA} \\ \cline{1-8}
Dataset                  & Attack & Bucket                                                 & GAS                                                   & Bucket                                                 & GAS                                                    & Bucket                                                 & GAS                                                    &                        \\ \hline
\multirow{2}{*}{CIFAR10} & LIE    & \begin{tabular}[c]{@{}c@{}}80.57\\ +0.67\end{tabular}  & \begin{tabular}[c]{@{}c@{}}80.86\\ +0.96\end{tabular} & \begin{tabular}[c]{@{}c@{}}79.43\\ +2.19\end{tabular}  & \begin{tabular}[c]{@{}c@{}}80.00\\ +2.76\end{tabular}  & \begin{tabular}[c]{@{}c@{}}80.38\\ +2.38\end{tabular}  & \begin{tabular}[c]{@{}c@{}}80.19\\ +2.19\end{tabular}  & \textbf{84.95}                  \\ \cline{2-9} 
                         & IPM    & \begin{tabular}[c]{@{}c@{}}10.95\\ -71.43\end{tabular} & \begin{tabular}[c]{@{}c@{}}82.38\\ +0\end{tabular}    & \begin{tabular}[c]{@{}c@{}}10.95\\ -64.10\end{tabular} & \begin{tabular}[c]{@{}c@{}}82.76\\ +7.71\end{tabular}  & \begin{tabular}[c]{@{}c@{}}10.95\\ -25.53\end{tabular} & \begin{tabular}[c]{@{}c@{}}82.76\\ +46.28\end{tabular} & \textbf{84.29}                  \\ \hline
\multirow{2}{*}{EMNIST}  & LIE    & \begin{tabular}[c]{@{}c@{}}25.01\\ +22.14\end{tabular} & \begin{tabular}[c]{@{}c@{}}2.87\\ +0\end{tabular}     & \begin{tabular}[c]{@{}c@{}}27.31\\ +24.44\end{tabular} & \begin{tabular}[c]{@{}c@{}}2.87\\ +0\end{tabular}      & \begin{tabular}[c]{@{}c@{}}30.28\\ +27.41\end{tabular} & \begin{tabular}[c]{@{}c@{}}2.87\\ +0\end{tabular}      & \textbf{85.16}                  \\ \cline{2-9} 
                         & IPM    & \begin{tabular}[c]{@{}c@{}}10.60\\ +8.54\end{tabular}  & \begin{tabular}[c]{@{}c@{}}3.96\\ +1.90\end{tabular}  & \begin{tabular}[c]{@{}c@{}}10.60\\ -55.28\end{tabular} & \begin{tabular}[c]{@{}c@{}}4.02\\ -61.86\end{tabular}  & \begin{tabular}[c]{@{}c@{}}10.60\\ -8.46\end{tabular}  & \begin{tabular}[c]{@{}c@{}}2.87\\ -16.19\end{tabular}  & \textbf{85.43}                  \\ \hline
\multirow{2}{*}{WISDM}   & LIE    & \begin{tabular}[c]{@{}c@{}}93.57\\ -0.99\end{tabular}  & \begin{tabular}[c]{@{}c@{}}94.71\\ +0.15\end{tabular} & \begin{tabular}[c]{@{}c@{}}93.19\\ +0.44\end{tabular}  & \begin{tabular}[c]{@{}c@{}}94.14\\ +1.39\end{tabular}  & \begin{tabular}[c]{@{}c@{}}94.26\\ +1.71\end{tabular}  & \begin{tabular}[c]{@{}c@{}}94.86\\ +2.31\end{tabular}  & \textbf{94.56}                  \\ \cline{2-9} 
                         & IPM    & \begin{tabular}[c]{@{}c@{}}38.94\\ -54.41\end{tabular} & \begin{tabular}[c]{@{}c@{}}91.49\\ -1.86\end{tabular} & \begin{tabular}[c]{@{}c@{}}38.94\\ -42.73\end{tabular} & \begin{tabular}[c]{@{}c@{}}92.06\\ +10.39\end{tabular} & \begin{tabular}[c]{@{}c@{}}37.36\\ -45.02\end{tabular} & \begin{tabular}[c]{@{}c@{}}93.55\\ +11.17\end{tabular} & \textbf{94.06}                  \\ \hline
\end{tabular}
\vspace{2mm}
\end{table}
\vspace{2mm}
\subsubsection{AGRs Augmented with Non-IID Denfenses}\label{sec:agraug}

In Section~\ref{sec:rw}, we introduced two SOTA non-IID defenses (i.e., Bucketing and GAS). To evaluate the robustness of combining non-IID defenses with AGRs in non-IID settings, we compare the performance of M-Krum, Median, and RFA when combined with Bucketing and GAS using three datasets. Table~\ref{tab:agraug} presents the model performance and the performance improvement after combination, with negative numbers indicating a performance decline. From Table~\ref{tab:agraug}, it is evident that under the LIE attack, both non-IID defenses lead to performance improvements, especially on EMNIST. However, under the IPM attack, combining non-IID defenses with AGRs even brings negative effects on model performance, particularly with the Bucketing method. The reason is that before using AGRs for inter-bucket aggregation, average aggregation has occurred in each bucket. This two-step aggregation might cause overly aggressive cancellation, resulting in too many beneficial model updates being excluded from model aggregation, ultimately affecting the model performance. These results indicate that the incorporation of non-IID defenses still cannot generalize across various attacks to assist AGRs in enhancing the robustness.

In contrast, FedCAP withstands attacks in all cases due to its model update calibration mechanism, which enables the server to differentiate malicious model updates from benign ones in non-IID settings. With the enhancement of customized aggregation, FedCAP accelerates the deterioration of malicious client models, ensuring that the anomaly detection mechanism quickly identifies and removes malicious clients, effectively defending against attacks.
% \vspace{-1mm}
\subsection{Robustness of FedCAP}

\begin{table}[h]
\vspace{-3mm}
  \caption{Robustness Analysis of FedCAP}
  \label{tab:FPNR}
  \centering
  \scriptsize
  \setlength{\tabcolsep}{3pt}
\begin{tabular}{c|c|ccccccc}
\hline
Dataset  & Metrics                                                         & Benign                                                    & SF                                                          & MR                                                          & LIE                                                             & Min-Max                                                        & Min-Sum                                                        & IPM                                                           \\ \hline
CIFAR-10 & \begin{tabular}[c]{@{}c@{}}TAcc\\ DAcc\\ FPR\\ FNR\end{tabular} & \begin{tabular}[c]{@{}c@{}}85.60\\ -\\ -\\ -\end{tabular} & \begin{tabular}[c]{@{}c@{}}84.00\\ 100\\ 0\\ 0\end{tabular} & \begin{tabular}[c]{@{}c@{}}84.10\\ 100\\ 0\\ 0\end{tabular} & \begin{tabular}[c]{@{}c@{}}84.95\\ 70\\ 0\\ 100\end{tabular}    & \begin{tabular}[c]{@{}c@{}}84.00\\ 100\\ 0\\ 0\end{tabular}    & \begin{tabular}[c]{@{}c@{}}84.38\\ 100\\ 0\\ 0\end{tabular}    & \begin{tabular}[c]{@{}c@{}}84.29\\ 100\\ 0\\ 0\end{tabular}   \\ \hline
EMNIST   & \begin{tabular}[c]{@{}c@{}}TAcc\\ DAcc\\ FPR\\ FNR\end{tabular} & \begin{tabular}[c]{@{}c@{}}86.59\\ -\\ -\\ -\end{tabular} & \begin{tabular}[c]{@{}c@{}}85.24\\ 80\\ 0\\ 20\end{tabular} & \begin{tabular}[c]{@{}c@{}}85.82\\ 100\\ 0\\ 0\end{tabular} & \begin{tabular}[c]{@{}c@{}}85.16\\ 81\\ 0\\ 63.33\end{tabular}  & \begin{tabular}[c]{@{}c@{}}85.39\\ 93\\ 0\\ 23.33\end{tabular} & \begin{tabular}[c]{@{}c@{}}85.58\\ 96\\ 0\\ 13.33\end{tabular} & \begin{tabular}[c]{@{}c@{}}85.43\\ 98\\ 0\\ 6.67\end{tabular} \\ \hline
WISDM    & \begin{tabular}[c]{@{}c@{}}TAcc\\ DAcc\\ FPR\\ FNR\end{tabular} & \begin{tabular}[c]{@{}c@{}}93.78\\ -\\ -\\ -\end{tabular} & \begin{tabular}[c]{@{}c@{}}94.16\\ 100\\ 0\\ 0\end{tabular} & \begin{tabular}[c]{@{}c@{}}95.07\\ 100\\ 0\\ 0\end{tabular} & \begin{tabular}[c]{@{}c@{}}94.56\\ 72.22\\ 0\\ 100\end{tabular} & \begin{tabular}[c]{@{}c@{}}94.06\\ 100\\ 0\\ 0\end{tabular}    & \begin{tabular}[c]{@{}c@{}}94.26\\ 100\\ 0\\ 0\end{tabular}    & \begin{tabular}[c]{@{}c@{}}94.06\\ 100\\ 0\\ 0\end{tabular}   \\ \hline
\end{tabular}
\vspace{-1mm}
\end{table}
To further demonstrate the robustness of FedCAP, in addition to evaluating the impact of attacks on model performance (i.e., TAcc), we introduce three robustness metrics to measure the detection abilities of FedCAP. \textbf{Detection Accuracy (DAcc)} represents the proportion of clients that are correctly identified as either benign or malicious. \textbf{False Positive Rate (FPR)} (or \textbf{False Negative Rate (FNR)}) denotes the proportion of benign (or malicious) clients that are incorrectly regarded as malicious (or benign). As can be seen from Table~\ref{tab:FPNR}, in most cases, FedCAP can identify and remove almost all malicious clients without mistakenly identifying benign clients as malicious with the FPR = 0. 

In particular, FedCAP fails to identify malicious clients under the LIE attack, with the FNR = 100 on CIFAR-10 and WISDM. The reason is that the collaboration among malicious clients during customized aggregation does not exacerbate the deterioration of malicious models and the Euclidean norm of malicious model updates does not reach the detection threshold $T_{norm}$, thus not triggering the anomaly detection mechanism. Surprisingly, in this case, the model performance of FedCAP is not significantly affected by the attack, with accuracy loss averaging less than 1\% compared to benign scenarios. This is attributed to its ability to isolate malicious model updates from benign ones during customized aggregation, so the aggregated customized models of benign clients are not seriously affected by the attack. 

\subsection{Impact of Hyperparameters}
\vspace{-3mm}
\begin{figure}[h]
\centering
\includegraphics[width=0.42\textwidth]{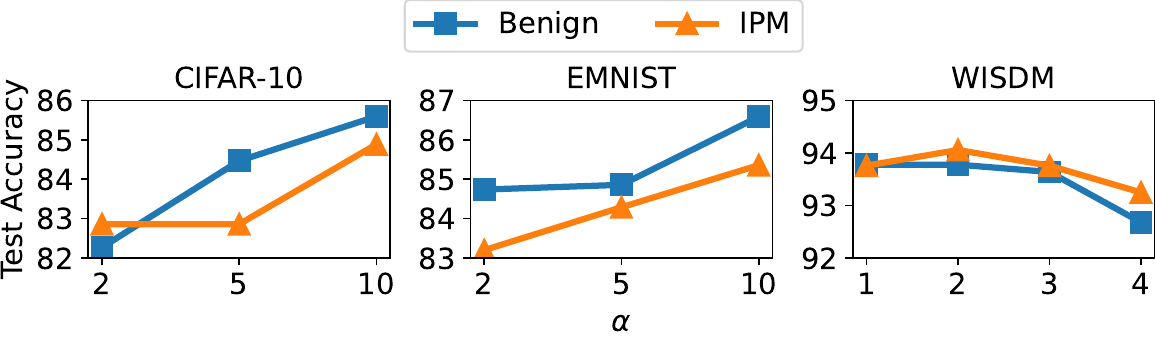}
  \vspace{-1mm}
  \caption{Impact of scale factor $\alpha$ 
}\label{fig:scale}
  % \vspace{-4mm}
\end{figure}

\begin{figure*}[h]
\centering
%\vspace{-4mm}
\includegraphics[width=0.98\textwidth]{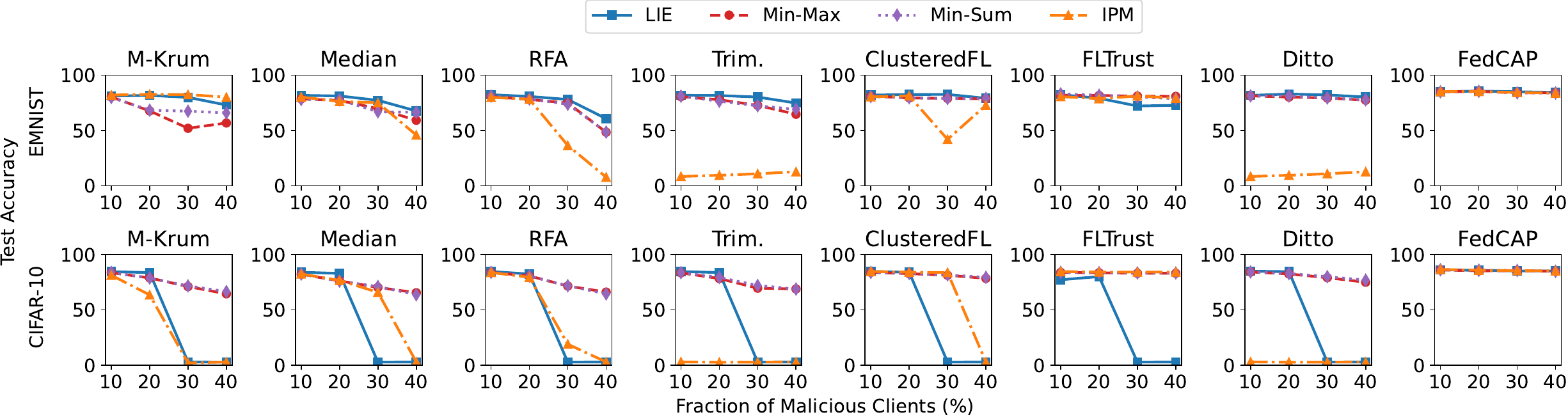}
  % \vspace{-2mm}
  \caption{Impact of the proportion of malicious clients
}\label{fig:atkR}
  \vspace{-4mm}
\end{figure*}

\subsubsection{Impact of Scale Factor $\alpha$}

To analyze the impact of $\alpha$ on the model performance of FedCAP, we conduct experiments using three datasets in both benign and attack scenarios. Here, we only discuss the parameter analysis under the IPM attack, as we find that the conclusions for other attacks are similar to it. As shown in Fig.~\ref{fig:scale}, for the WISDM dataset, as $\alpha$ increases, the model accuracy of FedCAP decreases. This result is attributed to the fact that there are many shared features among user data. For example, all users have highly similar behavior patterns such as walking. Therefore, when customizing aggregation weights, the server should choose a smaller value for $\alpha$ to fairly consider all users and learn more global features. In contrast, for the other two datasets, as $\alpha$ increases, the model accuracy of FedCAP gradually improves. The reason is that there is a higher degree of label distribution skew among clients' data. In this case, choosing a larger value for $\alpha$ allows the server to assign larger weights to clients with similar distributions, preventing interference from other dissimilar or malicious clients, and thus reducing the impact of data heterogeneity or attack. 

It is important to note that the choice of $\alpha$’s value depends on the degree of heterogeneity among client data distributions. In real-world attack scenarios, given the unknown number of malicious clients, FedCAP relies on its model update calibration and anomaly detection mechanisms to identify malicious clients, rather than tuning $\alpha$.

\subsubsection{Impact of Weight Factor $\phi$}\label{sec:phi}
\begin{figure}[h]
\centering
  \includegraphics[width=0.42\textwidth]{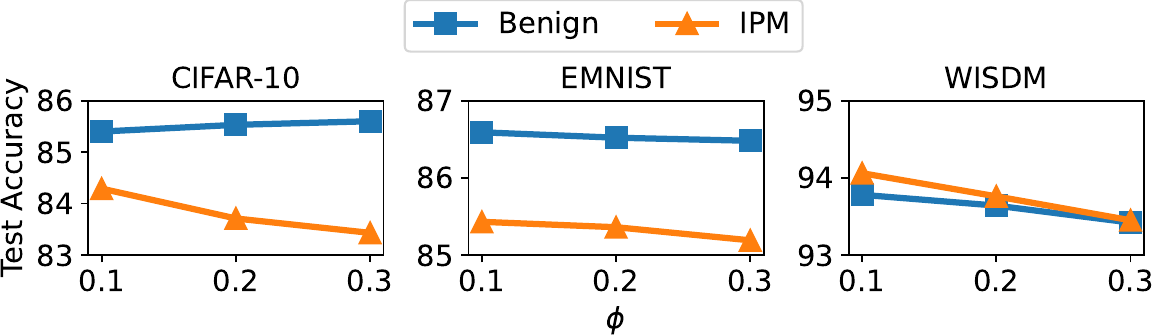}
  \vspace{-2mm}
  \caption{Impact of weight factor $\phi$ 
}\label{fig:phi}
  % \vspace{-4mm}
\end{figure}

To investigate the impact of $\phi$ on the model performance of FedCAP, we conduct experiments on three datasets in both benign and attack scenarios (e.g., IPM). As shown in Fig.~\ref{fig:phi}, the value of $\phi$ is inversely proportional to the model accuracy. The reason is that as $\phi$ increases, the weight corresponding to the client itself becomes larger, which weakens the reference to valuable knowledge from other client models in the model customization, thereby affecting the model performance. Additionally, we find that the performance of FedCAP is relatively robust to the choice of $\phi$'s value.

\subsubsection{Impact of Regularization Factor $\lambda$}
\begin{figure}[h]
\centering
\includegraphics[width=0.42\textwidth]{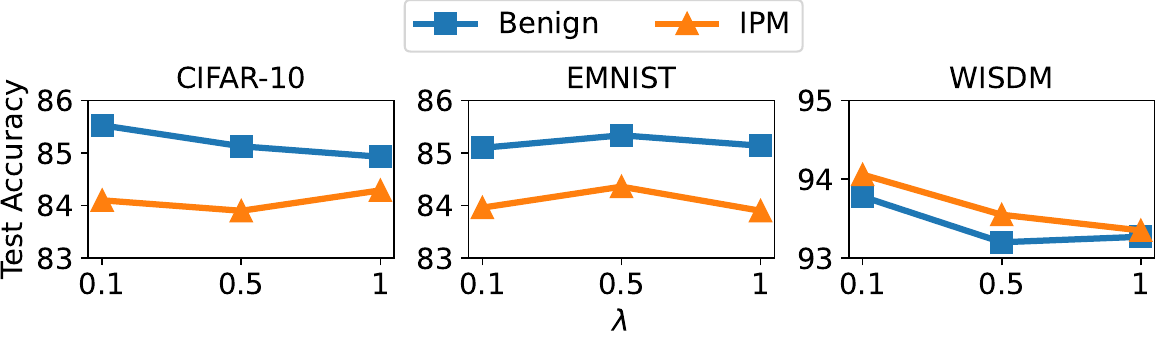}
  \vspace{-2mm}
  \caption{Impact of regularization factor $\lambda$ 
}\label{fig:lambda}
  \vspace{2mm}
\end{figure}

To explore the impact of $\lambda$ on model performance of FedCAP, we conduct experiments using three datasets in both benign and attack scenarios (e.g., IPM). As shown in Fig.~\ref{fig:lambda}, we conclude that appropriately increasing $\lambda$ not only allows the personalized model training to reference global knowledge from the customized model but also prevents overfitting of the personalized model on the limited data. When $\lambda$ is too large, the training of the personalized model is influenced by the customized model that may still be affected by data heterogeneity or attack, leading to a decrease in its accuracy.

\subsubsection{Impact of the Proportion of Malicious Clients}
To evaluate the impact of the proportion of malicious clients on the effectiveness of robust FL methods, we use the CIFAR-10 and EMNIST datasets to conduct experiments under four strong attacks (i.e., LIE, Min-Max, Min-Sum, and IPM). Fig.~\ref{fig:atkR} shows that as the proportion of malicious clients increases, especially to 30\% and 40\%, the model performance of most robust FL baselines is significantly affected by attacks and decreases substantially. Moreover, although FLTrust and Ditto resist attacks to some extent on CIFAR-10, their effectiveness significantly is suppressed on EMNIST, indicating that they cannot demonstrate robustness across various attacks in non-IID settings. Conversely, regardless of the proportion of malicious clients, FedCAP exhibits strong robustness and generalizes well across various attacks while maintaining superior model performance.

\subsection{Ablation Study}\label{sec:Ablation_Study}

\begin{table}[h]
\vspace{-4mm}
  \caption{Ablation Study Analysis of FedCAP components}
  \label{tab:Ablation_Study}
  %\vspace{-3mm}
  \centering
  \scriptsize
  \setlength{\tabcolsep}{0.3pt}
\begin{tabular}{c|c|c|c|ccccc}
\hline
      & Cust. Agg. & Calibration & Pers. Train. & Benign & LIE   & Min-Max & Min-Sum & IPM   \\ \hline
\ding{172}      & \checkmark         &             &             & 83.53  & 78.67 & 48.76   & 56.29   & 83.14 \\
\ding{173}      &           &             & \checkmark           & 82.73  & 82.10 & 79.33   & 79.33   & 10.95 \\
\ding{174}      & \checkmark         & \checkmark           &             & 85.07  & 84.29 & 83.52   & 83.81   & 83.71 \\
\ding{175}      & \checkmark         &             & \checkmark           & 83.60  & 80.10 & 69.81   & 70.95   & 83.14 \\
\ding{176}(FedCAP) & \checkmark         & \checkmark           & \checkmark           & \textbf{85.60}  & \textbf{84.95} & \textbf{84.00}   & \textbf{84.38}   & \textbf{84.29} \\ \hline
\end{tabular}
 \vspace{1mm}
\end{table}
%To verify the effectiveness of the three main components
To verify the necessity of each main component in FedCAP (i.e., customized aggregation, model update calibration, and personalized training) as introduced in Section~\ref{sec:FedCAP_Sys}, we conduct ablation experiments using CIFAR-10 under four strong attacks and analyze the results in Table~\ref{tab:Ablation_Study}. 

First, \ding{172}'s model performance is about 2\% lower than \ding{174}’s in the benign scenario and about 17\% lower on average in attack scenarios. This indicates that without the model update calibration mechanism, the standalone customized aggregation mechanism cannot precisely capture the similarity relationships among benign clients, nor can it differentiate between malicious and benign model updates, leading to a significant drop in the performance of customized models due to the impact of data heterogeneity and attacks.

Second, \ding{173}’s model performance is about 3\% lower than \ding{176}’s in the benign scenario, and, while it shows some robustness to attacks other than IPM, the result indicates that the single personalized training mechanism still lacks the robustness to generalize across various attack scenarios.

Third, \ding{174}’s model performance is on average less than 1\% lower than \ding{176}'s, suggesting that the combination of customized aggregation and model calibration mechanisms already significantly mitigates the impact of data heterogeneity and attacks, and the addition of the personalized training further enhances model accuracy and robustness.

Last, \ding{175}’s model performance is 2\% lower than \ding{176}’s in the benign scenario and about 8\% lower on average in attack scenarios. This result shows that simply combining customized aggregation with personalized training mechanisms cannot achieve superior model performance. The incorporation of the model update calibration ensures malicious model updates are differentiated, which enables the server to isolate malicious clients from benign ones during customized aggregation, accelerate the deterioration of malicious models, and trigger the anomaly detection mechanism to identify and remove malicious clients permanently. Based on this, the proposed personalized training module can further improve robustness and model performance, building on the relatively clean customized models.

% First, by comparing \ding{172} with \ding{174}, we find that, without model update calibration, solely using customized aggregation fails to withstand many attacks. This deficiency stems from relying only on cosine similarity as the distance measurement, which cannot accurately distinguish between the model updates of malicious and benign clients. The Introduction of the update calibration module greatly contributes to revealing the differences between malicious and benign updates. Second, by comparing \ding{172} with \ding{175} and comparing \ding{174} with \ding{176}, we can confirm that the incorporation of personalized training can further mitigate the impact of the attacks (e.g., Min-Max/Min-Sum). However, the sole use of personalized training, as shown in \ding{173} fails to resist all attacks (e.g., IPM). 

Therefore, all the proposed modules in FedCAP are essential and contribute to achieving superior accuracy and robustness across various non-IID settings and attacks.
% \vspace{-1mm}

\section{Conclusion and Future Work}
% \vspace{-1.2mm}
In this paper, we proposed FedCAP,  a robust FL framework against both data heterogeneity and Byzantine attack. Specifically, we designed a customized model aggregation scheme, which can facilitate collaborative training among similar clients and accelerate the deterioration of malicious models. In addition, we developed a model update calibration mechanism to capture the differences in the direction and magnitude of model updates among clients and an anomaly detection mechanism to help the server quickly identify and permanently remove malicious clients. Extensive experiments demonstrated that FedCAP outperformed the SOTA baselines in terms of model accuracy under several non-IID settings and model robustness under various types of poisoning attacks.

\noindent\textbf{Robustness Analysis.} We believe that in strong attack scenarios~\cite{LIE, Shejwalkar2021ManipulatingTB, DBLP:conf/uai/XieKG19}, even if adversaries know the customized aggregation rule and the model updates of benign clients, it is difficult to design effective adaptive attacks. The reasons are twofold: first, the model updates uploaded by clients are calibrated by the server, making the knowledge adversaries possess about the uploaded model updates outdated. Second, the server aggregates the customized model for each client, making it challenging for adversaries to manipulate malicious model updates that can affect the model customization of all benign clients. In future work, we will conduct more theoretical robustness analysis of FedCAP and explore potential adaptive attack strategies.

\noindent\textbf{Convergence Analysis.} In FedCAP, the server’s customized aggregation and model update calibration mechanisms dynamically adjust the aggregation process based on incoming updates. This dynamic nature introduces additional layers of complexity for theoretical analysis. In future work, we can decompose the convergence analysis into multiple aspects, such as studying the impact of varying customized aggregation weights on model convergence speed and final performance, theoretically deriving how the model update calibration mechanism affects model convergence, and investigating the impact of regularization factors in the personalized objective function on personalized model training.

\section*{Acknowledgment}

In this work, Lichao Sun was partially supported by the National Science Foundation (NSF) (CRII-2246067, ATD-2427915, NSF POSE-2346158, and Lehigh Grant FRGS00011497). Xuyu Wang was supported in part by the NSF (CNS-2319343, CNS-2317190, CNS-2415209, and IIS-2306791).

\bibliographystyle{IEEEtran}
\bibliography{reference.bib}

\appendices

\section{System Scalability, Overhead, and Efficiency}\label{sec:sys_overhead}
\subsection{Scalability} In read-world deployments, FedCAP not only customizes models to meet participants' personalized needs but is also scalable for future clients. For example, when a future client $i$ requests the model customization service, the server sends the global model $\boldsymbol{w}^{T-1}$ to client $i$. Upon receiving the model, client $i$ conducts local updating and uploads the model update $\boldsymbol{d}_i$ to the server. Afterward, the server customizes the model $\boldsymbol{\hat{w}}_i$ using the uploaded model update $\boldsymbol{d}_i$, the calibrated update pool $\{\boldsymbol{\tilde{d}}^{T-1}_k\}_{k\in S_{T-1}}$, and the recovered model pool $\{\boldsymbol{\tilde{w}}^{T-1}_k\}_{k\in S_{T-1}}$ through Section~\ref{sec:customization}. Once client $i$ receives the customized model $\boldsymbol{\hat{w}}_i$, it can directly perform model inference or further fine-tuning. 

\subsection{Overhead} FedCAP proposes three modules: model customization, personalized training, and model update calibration. 

\subsubsection{Computational Overhead} During customized aggregation, the server only needs to compute cosine similarity among model updates of $N$ participating clients to determine the customized aggregation weights, where typically $N << K$ (the total number of clients). In the personalized training phase, although clients need to alternately update personalized models and customized models within the same mini-batch, results from ablation experiments in Section~\ref{sec:Ablation_Study} suggest that customized models already achieve satisfactory model performance. In the model update calibration phase, the computational overhead introduced is negligible compared to model customization. Although the computational complexity of existing AGRs (e.g., Median and RFA) can be reduced to $O(K)$, their combination with non-IID defenses (i.e., Bucketing and GAS) or personalized FL methods (e.g., Ditto) still cannot ensure robustness across various attack scenarios in non-IID settings (see Section~\ref{sec:agraug} and Section~\ref{sec:Ditto+}). Our work balances system overhead with robustness against both data heterogeneity and Byzantine attacks. Additionally, FedCAP is orthogonal to computation-efficient methods. For instance, since the classification layer of the model contains more personalized features~\cite{DBLP:conf/iclr/ChenC22, oh2022fedbabu}, future work will explore calculating similarity only for the last layer.

\subsubsection{Communication Overhead} Compared to the client-side model customization method FedFomo, where the server needs to send multiple models to clients in each communication round, FedCAP only needs to send one customized model to each client, resulting in communication overhead equivalent to FedAvg.

\subsubsection{Storage Overhead} FedCAP only requires storing model pools and model update pools for the $N$ participating clients. In real-world scenarios, central servers often have abundant storage resources~\cite{kairouz2021advances}, making storage overhead acceptable. Additionally, FedCAP can be combined with storage-efficient methods such as model quantization to further improve system efficiency.

In summary, considering the improvement in robustness against both data heterogeneity and Byzantine attacks provided by FedCAP, as well as the unsatisfactory performance improvement of the combination of robust FL and personalized FL methods, the introduced overhead is acceptable.

\subsection{Efficiency Analysis}

\begin{table}[h]
\vspace{-4mm}
  \caption{System Efficiency Analysis of FedCAP}
  \label{tab:efficiency}
\begin{threeparttable}
  \centering
  \scriptsize
  \setlength{\tabcolsep}{0.6pt}
\begin{tabular}{c|c|c|c|c}
\hline
Method    & R2Acc(80\%) & Computation(clients) & Computation(server) & Communication \\ \hline
FedAvg-FT & 52th         & 3.2min                & 0.2s\;\,                 & \;\,12MB / round  \\ \hline
FedRoD    & 34th         & 4.0min                & 0.2s\;\,                 & \;\,12MB / round  \\ \hline
FedFomo   & 11th         & 2.6min                & 0.9s\;\,                 & 120MB / round \\ \hline
Ditto     & 44th         & 6.2min                & 0.2s\;\,                 & \;\,12MB / round  \\ \hline
FedCAP    & \;\,9th          & 6.2min                & 2.6s*                 & \;\,12MB / round  \\ \hline
\end{tabular}
\begin{tablenotes}[flushleft] 
% \vspace{-0.02in}
\item *customized aggregation: 2.47s, calibration: 0.10s, detection: 0.07s
\end{tablenotes}
\end{threeparttable}
\end{table}

To compare FedCAP's system efficiency with other baselines, we report their Round-to-Accuracy (R2Acc), breakdown of computation time on both server and client sides, and communication overhead in the benign scenario using CIFAR-10. The R2Acc represents the number of rounds required to achieve a target accuracy (i.e., 80\% on CIFAR-10), which reflects the convergence speed of FL algorithms. The communication overhead involves the data volume of models transmitted back and forth between the server and clients, and its value is determined by the number and size of models transmitted each round.

As shown in Table~\ref{tab:efficiency}, FedCAP achieves the target accuracy in the 9th round, which is on average 4 times that of others. Although FedFomo shows comparable R2Acc (11th round), its communication overhead is about 10 times that of FedCAP, as its server needs to send multiple models (10 for CIFAR-10) to each client per round to compute aggregation weights. The client-side computation times of FedRoD, Ditto, and FedCAP are about 1.3, 1.9, and 1.9 times that of FedAvg+FT, respectively, because clients need to perform additional computations for training personalized heads or models. Moreover, although FedCAP's server-side components such as customized aggregation consume additional time compared to average aggregation, this is acceptable since FedCAP's R2Acc is significantly better than that of the others. This indicates that FedCAP can reduce the overall computational overhead by decreasing the number of FL training rounds (due to its high convergence speed), thereby improving system efficiency.

In terms of scalability, FedCAP has the same communication overhead as FedAvg, increasing with the complexity of the transmitted model. As the dataset size grows, although FedCAP's personalized training introduces additional computation overhead compared to FedAvg, its customized models have achieved high accuracy (see Table~\ref{tab:Ablation_Study}-\ding{174}, \ding{176}). Therefore, in read-world deployments, a trade-off between model performance and system efficiency can be achieved by selecting between customized or personalized models.

\end{document}